\documentclass[3p]{elsarticle}
\usepackage{hyperref}
\usepackage{float}
\usepackage{verbatim} 
\usepackage{apalike}
\usepackage{xcolor}
\restylefloat{figure}
\restylefloat{table}
\usepackage{amsmath}
\usepackage{algorithmic}
\usepackage{array}
\usepackage{fixltx2e}
\usepackage{stfloats}
\usepackage{amsfonts}
\usepackage{multicol}
\usepackage{float}
\usepackage{booktabs}
\usepackage{soul}

\newcommand{\ra}[1]{\renewcommand{\arraystretch}{#1}}

\journal{/ to be submitted}

\biboptions{authoryear}

\begin{document}
\begin{frontmatter}

\title{Antifragile Control Systems: The case of mobile robot trajectory tracking in the presence of uncertainty}

\author[label1]{Cristian~Axenie\corref{cor1}}
\ead{cristian.axenie@gmail.com}

\cortext[cor1]{Corresponding author.}
\address[label1]{Audi Konfuzius-Institut Ingolstadt Laboratory, Technische Hochschule Ingolstadt, Esplanade 10, 85049 Ingolstadt, Germany}
\author[label3]{Matteo Saveriano}
\ead{matteo.saveriano@unitn.it}
\address[label3]{Department of Industrial Engineering, University of Trento, Via Sommarive 9, 38123 Trento, Italy}

\begin{abstract}
Mobile robots are ubiquitous. Such vehicles benefit from well-designed and calibrated control algorithms ensuring their task execution under precise uncertainty bounds. Yet, in tasks involving humans in the loop, such as elderly or mobility impaired, the problem takes a new dimension. In such cases, the system needs not only to compensate for uncertainty and volatility in its operation but at the same time to anticipate and offer responses that go beyond robust. Such robots operate in cluttered, complex environments, akin to human residences, and need to face during their operation sensor and, even, actuator faults, and still operate. This is where our thesis comes into the foreground. We propose a new control design framework based on the principles of antifragility. Such a design is meant to offer a high uncertainty anticipation given previous exposure to failures and faults, and exploit this anticipation capacity to provide performance beyond robust. In the current instantiation of antifragile control applied to mobile robot trajectory tracking, we provide controller design steps, the analysis of performance under parametrizable uncertainty and faults, as well as an extended comparative evaluation against state-of-the-art controllers. We believe in the potential antifragile control has in achieving closed-loop performance in the face of uncertainty and volatility by using its exposures to uncertainty to increase its capacity to anticipate and compensate for such events.

\end{abstract}


\end{frontmatter}

\section{Introduction}

There are around 90 million elderly or handicapped persons in the European Union (EU). Even more, in 2019 alone, almost half of the EU elderly population (people aged 65 and over) reported difficulties with at least one personal care or household activity, according to EU Committee statistics from \cite{EUstats}. Various reports also demonstrate a close relationship between a person's age and the handicaps suffered, with the latter being more frequent in those of senior age. Given the EU's rising life expectancy, this indicates that a considerable proportion of its population will have functional issues. Recognizing the scarcity of applications for this segment of the population, governments and public institutions have been encouraging research in this area in recent years.

On a global scale, several research organizations have begun to establish cooperative projects, initiatives to improve communication and mobility of the elderly and/or disabled, with the goal of improving their quality of life and giving them a more autonomous and independent lifestyle. Targeting increased possibilities of social inclusion, the most recent initiative being the Bavarian Geriatronics Lighthouse Project of \cite{haddadin2018tactile} in Germany stands out as a leading example. Within this context, wheelchair deployment is straightforward and one of the most potentially beneficial applications for boosting the mobility of disabled and/or elderly people. A typical motorized wheelchair facilitates the movement of disabled persons who are unable to walk, provided that their impairment enables them to properly handle the joystick. 

Yet, individuals with severe disabilities or handicaps, on the other hand, may find it difficult or impossible to utilize them; for example, paraplegics, or even tetraplegics, who can only handle an on-off sensor or make certain extremely limited motions. This would make controlling the wheelchair challenging, especially during precise maneuvers. In such circumstances, more advanced human-wheelchair interfaces tailored to the user's impairment are required, allowing them to enter movement commands in a safe and straightforward manner. 
Robotic wheelchairs are the most straightforward alternative, that will accept the user's limited input, plan a trajectory, and travel along it within the task's time limits and the operational environment's complexity. This is the core focus of our study, how can we design a motion control algorithm, that allows safe motion in unstructured and uncertain environments, in the presence of uncertainty and volatility.

\subsection{Trajectory tracking for wheeled mobile robots}

Mobile robotics has sparked the control community's interest in the context of human-assistive applications. Such wheeled mobile robots are often characterized as nonholonomic mechanical systems. For many years, nonholonomic vehicle control has been a hotly debated research topic. This is due to at least two factors. On the one hand, nonholonomic wheeled robotic vehicles are an important and increasingly common mode of mobility. Previously only seen in research labs and factories, autonomous robotic vehicles are increasingly being adopted in everyday life (e.g. through car-platooning applications, as shown in autonomous mobility by \cite{althoff2022}, geriatronics applications of \cite{haddadin2020}, or urban transportation services described by \cite{chong2018simulation}). 

Trajectory tracking control of nonholonomic mobile robots seeks to control a robot's motion in order to follow a specific time-varying trajectory. It is a basic motion control challenge that the robotics community has studied in great detail, proof of the pioneering works of \cite{luca1995modelling}, \cite{oriolo2002wmr}, \cite{zhang2003variable}, \cite{solea2007trajectory}, and, of course, our previous work in \cite{axenie2010real}. The tracking control problem is categorized as either kinematic or dynamic depending on whether the system is described by a kinematic or dynamic model. 

Several researchers have investigated the kinematic tracking problem and developed various types of controllers. The seminal work of \cite{kanayama1990stable} addressed the trajectory-tracking problem by using the kinematic model of a wheeled mobile robot. In a more theoretical work \cite{jiangdagger1997tracking} addressed both local and global tracking problems with exponential convergence utilizing time variable state feedback based on the backstepping approach. As we see, the kinematic tracking control problem for mobile robots has received much research, but the dynamic tracking control problem has only recently attracted attention. 

The majority of the results on dynamic model-based tracking problems of nonholonomic systems are presented on the assumption that the system's kinematics are precisely understood and that uncertainties exist only in the dynamics. In practice, however, errors exist in both kinematics and dynamics. Typically, the reference trajectory is derived by employing a reference (virtual) robot; hence, the reference trajectory takes into account all kinematic restrictions implicitly. The majority of the control inputs are produced using a mix of feed-forward inputs estimated from the reference trajectory and feedback control rule, as shown in the work of \cite{luca1995modelling}, \cite{oriolo2002wmr}, and \cite{sarkar1994control}. In the same context, the work of \cite{samson1991feedback}, \cite{kanayama1990stable}, and, of course, \cite{samson1991feedback} pioneered Lyapunov stable time-varying state-tracking control rules, in which the system equations are linearized with regard to the reference. The controller parameters are calculated by defining the desired parameters of the characteristic polynomial. A nonzero motion condition is required for stability to the reference trajectory. Along the same lines, the work of \cite{zhang1997discontinuous} introduced a discontinuous stabilizing controller for mobile robots with nonholonomic restrictions, in which the robot's state asymptotically converges to the goal configuration with a smooth trajectory. 

More in line with our uncertainty handling approach, the work of \cite{koh1999smooth} developed a tracking problem for a mobile robot to follow a virtual target vehicle that moves precisely along a path with a given velocity. In order to minimize wheel slippage or mechanical damage during navigation, the driving velocity control rule was created based on bang-bang control while taking the acceleration boundaries of the driving wheels and the robot's dynamic restrictions into account. Also relevant, the work of \cite{zhang2003variable} designed a tracking controller for a differential drive mobile robot that is sensitive to wheel slip and external stresses using dynamic modeling.

As we see, many researchers have employed various nonlinear control strategies when dealing with system disturbances, operating uncertainty, and unknown dynamic characteristics. Similar in nature to one component of our approach, and used to tackle the tracking control problem for mobile robots, the pioneering work of \cite{5276210}, the work of \cite{yang1999sliding}, \cite{kim2000design}, \cite{fukao2000adaptive}, and \cite{li2002output} employed sliding mode motion control techniques, robust adaptive control techniques, and higher order sliding mode techniques, respectively. More precisely, they proposed variable structure control approaches, using sliding mode control for the trajectory tracking issue for mobile robots in the presence of disturbances that violate the nonholonomic constraints. Finally, \cite{wu2001path} and the excellent work of \cite{jiang2001saturated} established a model-based control design technique for the kinematic model with a nonholonomic mobile robot in the presence of input saturations that deals with global stabilization and global tracking control which yielded comparable results to the non-parametric adaptive control approach of \cite{pourboghrat2002adaptive} and the neural network robust control approach of \cite{dong2005robust}.

Although well rooted in the robotic control field, among the previously presented works, the current study covers an unique control approach for nonholonomic vehicles, namely antifragile control, and more precisely trajectory tracking in the face of uncertainty, volatility, and unpredictability. This approach goes beyond our initial explorations in \cite{axenie2010adaptive}, and tries to propose to the community a novel perspective on robot motion control, namely antifragile control.

\subsection{Fragility-robustness-antifragility spectrum in robot control}

Trajectory tracking requires a task planning step. At the planning level, autonomous robot vehicles produce their own judgments that determine how to operate the vehicle actuators and cause the vehicle to move, as shown in the work of \cite{8793786}. The challenge with motion planning and control is that the motion constraints of any actuators involved or the vehicle platform itself must be considered, as formally described in \cite{8967981}. This is especially relevant for wheeled mobile robots, which are constrained by nonholonomic constraints. This means that a vehicle traveling on a surface may have three degrees of freedom: two degrees of translation and one degree of rotation. As a result, the equations of motion that describe vehicle dynamics are non--integrable, making the problem significantly more complex to solve. This also implies that wheeled mobile robots are underactuated. In other terms, the system's number of control inputs is smaller than the number of degrees of freedom in its configuration space. Additionally, the uncertainty related to the traveling surface, sensors, and actuator faults is an additional dimension to consider in control design.

The main goal of this study is to introduce the application of antifragile control to mobile robot trajectory tracking control under uncertainty, volatility, and variability of the operating environment and the robot's sensors and actuators. According to \cite{taleb2012antifragile}, antifragility is a feature of a system that allows it to benefit from uncertainty, unpredictability, and volatility, in contrast to fragility. The reaction of an antifragile system to external perturbations is beyond robust and resilient such that mild stresses can increase the system's future response by adding a significant anticipation component. In this work, we propose an alternative control mechanism, based on the antifragile control framework introduced by \cite{axenie2022antifragile}, further refined and extended in \cite{axenie2022antifragiletraffic} and built on top of the principles in the seminal work of \cite{taleb2013mathematical}.

In order to instantiate the antifragile control framework for robot trajectory tracking control, we need to define the Fragility-robustness-antifragility spectrum. In order to guide the reader with an intuition on the benefits of antifragile control, we consider a simple depiction of how various types of controllers would perform in the presence of gradually increasing uncertainty (e.g. wheel slippage, actuator fault, or sensor fault). We consider a hypothetical effect only for graphical purposes.

\begin{figure}[H]
\includegraphics[scale=0.2]{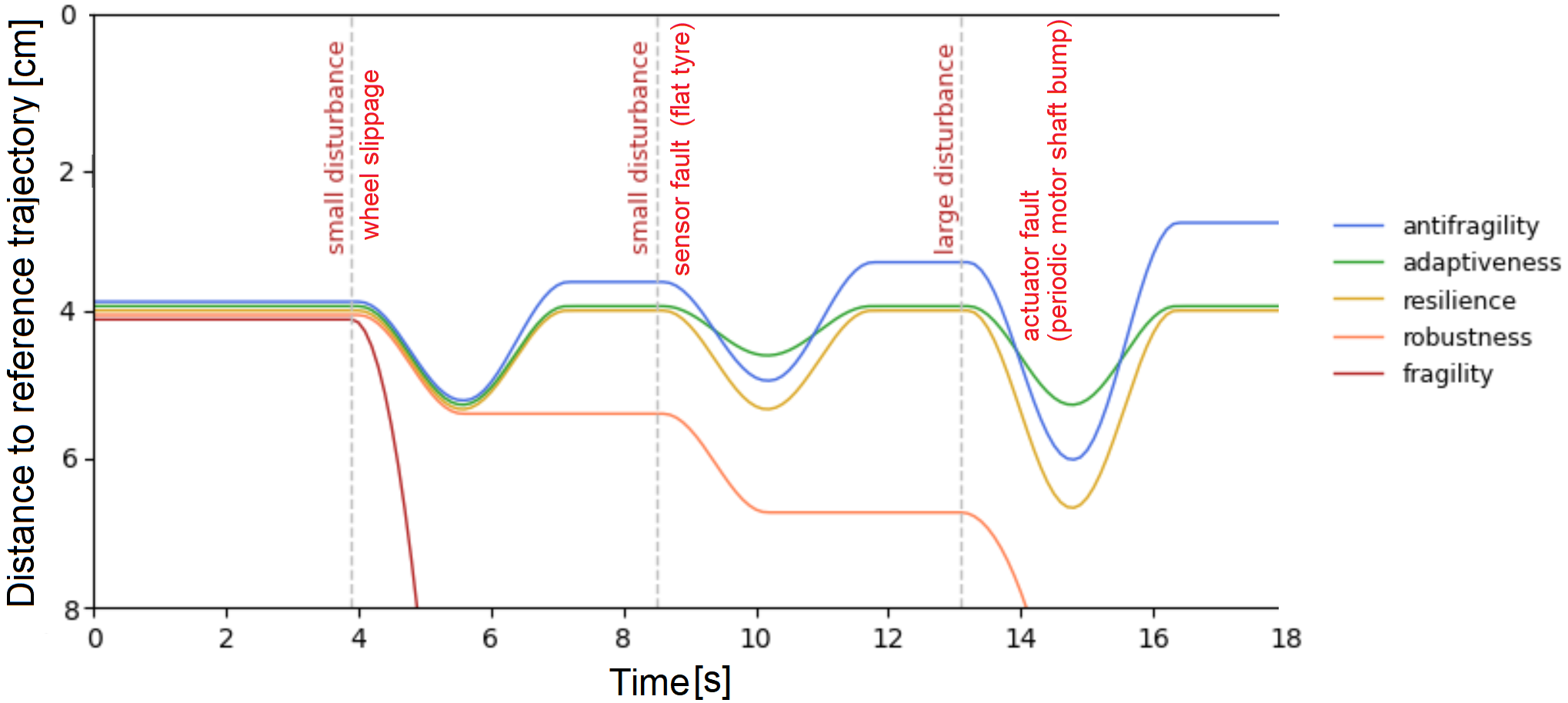}
\centering
\caption{Fragility-robustness-antifragility spectrum in robot trajectory tracking control. Uncertainty in a robot's motion can emerge from environmental parameters (e.g. wheel slippage), sensor faults (e.g. perceiving a continuous wheel radius decrease during operation akin to a flat tire), or actuator faults (e.g. a periodic eccentric mechanical motion of the DC motor shaft akin to a wheel bump). The possible closed-loop system responses are reflected in the actual displacement from the reference trajectory to track. We can clearly see that with the increase in amplitude and timing (i.e. uncertainty and volatility) of the disturbance the system can compensate up to a point but, in contrast to antifragile control, cannot gain from the adverse events. It is important to note the reaction time and the amplitude of the response with respect to the occurrence and strength of the adverse event.}
\label{fig1}
\end{figure}

The purpose of Figure~\ref{fig1} is to delineate, in a graphical and easy-to-grasp manner, the main concept of the proposed approach. Achieving an antifragile closed-loop control performance, that not only compensates for unexpected, increasingly strong disturbances but also gains from subsequent exposures, is the core motivation of our work. The actual implementation details follow in the next sections along with more intuitive aspects that strengthen this hypothetical depiction of the robot's response.

An important final note is that, in the current study, we extend the intrinsic and inherited fragility--robustness--antifragility detection heuristics of \cite{taleb2013mathematical} through a novel type, termed induced antifragility. Basically, we propose realizing induced antifragility through a design of a closed-loop control system that can judiciously compute motion control signals of the robot that compensate for uncertainty and volatility during trajectory tracking.

\subsection{Contributions}

The major contribution of this paper revolves mostly around another instantiation of the unique framework of antifragile control for mobile robot trajectory tracking control. Designing and implementing such a closed-loop control system requires both a good understanding of the system's dynamics, and the control task, and, of course, mapping the fragility-robustness-antifragility spectrum onto the design process. The main contributions of our study, and highlights of the following sections, are:

\begin{itemize}
\item a systematic characterization of mobile robot trajectory tracking problem under uncertainty and volatility (i.e. environment conditions, sensor and actuator faults in a space--time--intensity reference system);
\item a control system design mapping the mobile robot trajectory tracking problem under sensor and actuator faults to the fragile--robust--antifragile continuum;
\item a control system synthesis method;
\item an implementation of a mobile robot trajectory tracking antifragile controller with closed-loop benefits from variability and volatility;
\item an evaluation and discussion of our results on a suite of simulated experiments.
\end{itemize}

\section{Materials and Methods}

In this section, we introduce the models and tools we employed in our study. We commence with a formal description of the robot control problem. We then delve directly into the system's analysis and design within the antifragile control framework. We conclude the section with the actual controller synthesis that was used in the experiments.

\subsection{Wheeled mobile robot trajectory tracking}

This subsection provides an overview of the modeling of nonholonomic mobile robots for trajectory tracking. We remind that in trajectory tracking mode, the real mobile robot must track a virtual mobile robot's trajectory under time constraints (see figure~\ref{fig4}). The motion control of such robots is subject to nonholonomic constraints, making motion perpendicular to the wheels impossible. Although the complete robot state must be measured, this constraint requires a nontrivial control mechanism. Because trajectory tracking is comparable to servosystems, it is ensured that the system will converge to the intended trajectory in deterministic time using an asymptotically stable control law (save for the perturbations that it may experience).

\begin{figure}[H]
\includegraphics[scale=0.28]{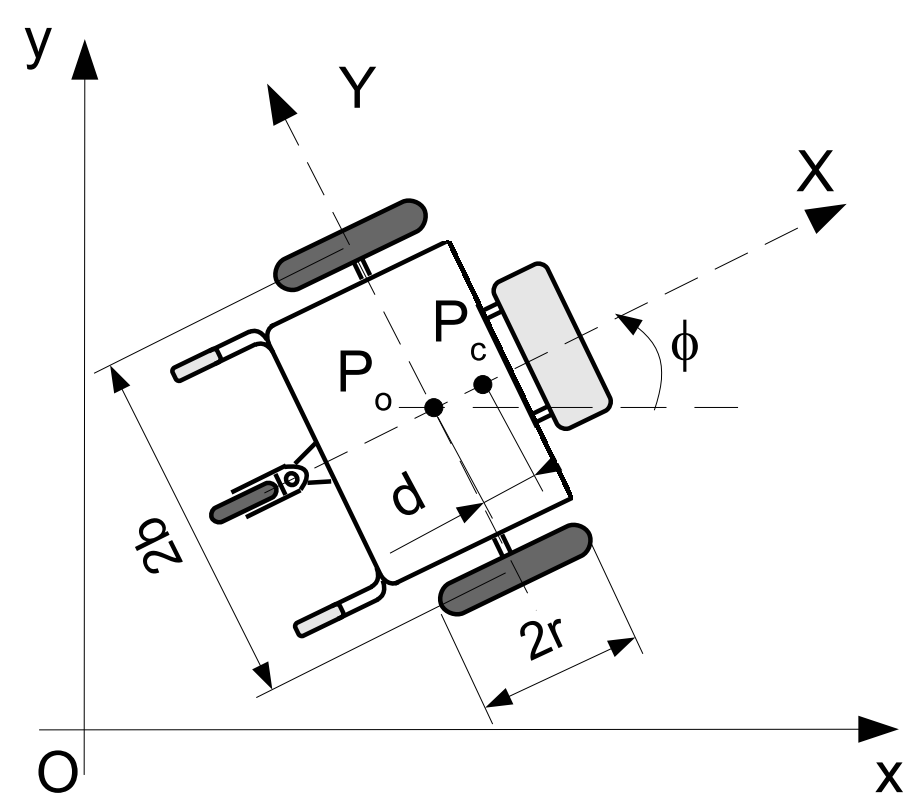}
\centering
\caption{Wheelchair type mobile robot with differential drive used in our study. All kinematic quantities are defined in the local coordinate (reference) system $XP_0Y$, whereas the control and measurements will be mapped to the world reference system $xOy$. $P_0$ is the origin of the local coordinate system fixed at the middle point between the right and left driving wheels. The distance from $P_0$ to the center of mass $P_c$ is $d$. Each driving wheel has a radius $r$ and the distance between wheels is $2b$. The heading angle of the robot is $\phi$. Adapted with permission from \cite{solea2009sliding}.}
\label{fig2}
\end{figure}

In our study, we consider a differential drive wheelchair as depicted in Figure~\ref{fig2}. The notations for the reference systems and the kinematic quantities follow the standard conventions. Additionally, for our robot, we assume that the velocity of $P_0$ must be in the direction of the axis of symmetry and the wheels must not skid (i.e. motion constraints).

\begin{figure}[H]
\includegraphics[scale=0.3]{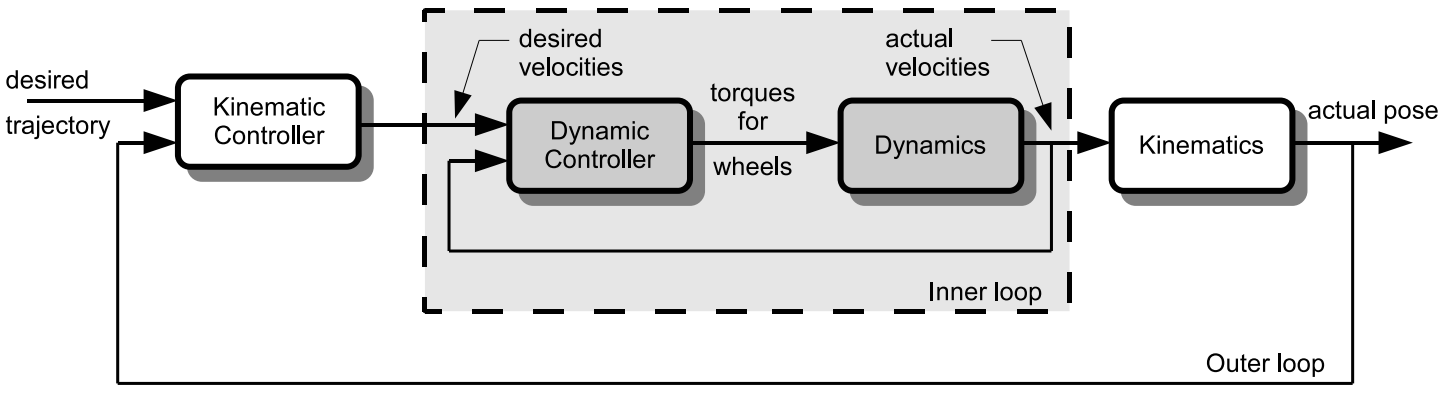}
\centering
\caption{Global control structure for mobile robot trajectory tracking. An inner loop is utilized to control both the linear and angular velocities. An outer loop to control the pose of the robot in the world coordinate system.}
\label{fig3}
\end{figure}

\begin{figure}[H]
\includegraphics[scale=0.3]{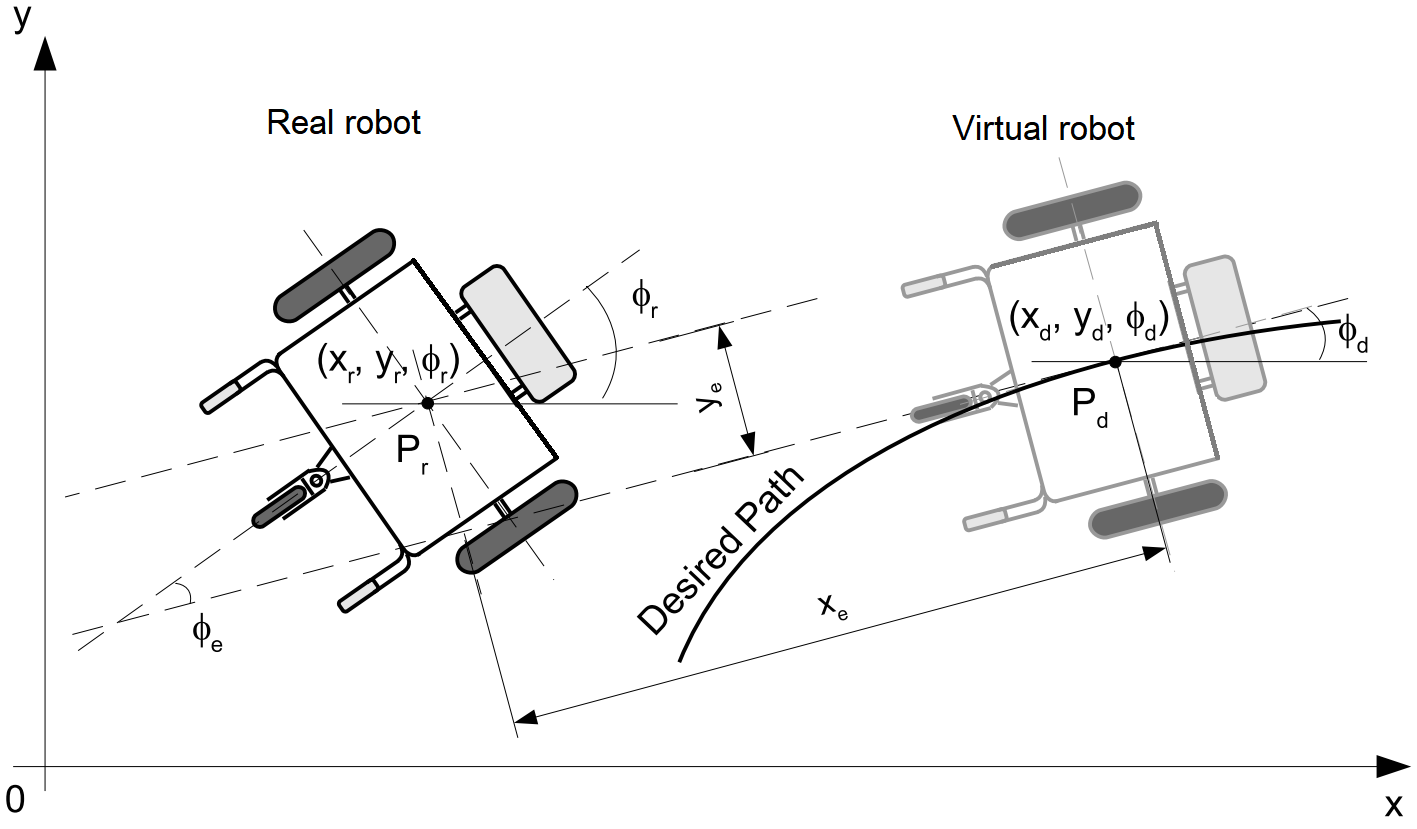}
\centering
\caption{Description of the mobile robot trajectory tracking problem. The real mobile robot tries to follow the desired path under time constraints. The control algorithm needs to compensate for the heading $\Phi_e$, lateral $y_e$, and longitudinal errors $x_e$ and come closer to the virtual robot. The goal is to make the robot pose $P_r = [x_r,y_r,\phi_r]^{\top}$ follow a reference trajectory of the virtual robot $P_d = [x_d,y_d,\phi_d]^{\top}$. Adapted with permission from \cite{solea2009sliding}.}
\label{fig4}
\end{figure}

In motion control, the objective is to control the velocity of the robot such that its pose $P = [x,y,\phi]^{\top}$ follows a reference trajectory. Initially, the study effort was centered solely on the kinematic model, assuming accurate velocity tracking. But this is not the case in real-world scenarios, where uncertainty and disturbances can make the closed-loop system unstable. In order to improve motion control performance, one must additionally consider individual vehicle dynamics. In this scenario, the controller structure should be divided into two phases, as shown in Figure~\ref{fig3}:
\begin{itemize}
\item an inner loop, depending on the robot dynamics, that can be utilized to control both the linear and angular velocities. It is also known as a mobile robot dynamic-level control.
\item  an outer loop to control the pose of the robot in the $xOy$ reference frame. It is also called kinematic-level control of a mobile robot.
\end{itemize}

A mobile robot system with an n-dimensional configuration space, generalized variables $(q_1,q_2,...,q_n)$, and constraints may be expressed, following the work of \cite{fierro1997control}, as following: 

\begin{equation}
M(q)\ddot{q} + V_m(q, \dot{q})\dot{q} + F(\dot{q}) + G(q) + \tau_d = B(q)\tau - A^{\top}(q)\lambda
\label{dynamics}
\end{equation}
where $M(q) \in R^{n\times n}$ is a symmetric positive definite inertia matrix of the robot, $V_m(q, \dot{q}) \in R^{n\times n}$ is the centripetal and Coriolis matrix, $F(\dot{q}) \in R^{n\times 1}$ describes the surface friction of the robot,  $G(q) \in R^{n\times 1}$ is the gravity vector, $\tau_d$ describes the overall bounded unknown disturbances including unstructured unmodelled dynamics, $B(q)\in R^{n\times r}$ is the input gain matrix, $\tau \in R^{n \times 1}$ is the input vector of the robot, $A(q)\in R^{m\times n}$ is the constraints matrix, and $\lambda \in R^{m \times 1}$ is the vector of constraint forces acting upon the robot.
The nonholonomic character of the mobile robot is associated with the notion that the robot's wheels roll without sliding. They are constrained by nonholonomic non--integrable equality requirements concerning velocity. In other words, the permissible velocity space has a lower dimension than the configuration space. This limitation can be expressed as $A(q)\dot{q} = 0$ where 

\begin{equation}
A(q) = 
\begin{bmatrix}
sin(\phi) & -cos(\phi) & d & 0  & 0 \\
cos(\phi) & sin(\phi)  & b & -r & 0 \\
cos(\phi) & sin(\phi)  &-b & 0  & -r
\end{bmatrix}
\label{constraints}
\end{equation}

But, for control, the configuration of the mobile robot may be described using five generalized coordinates, $q = [x,y,\phi,\theta_r,\theta_l]^{\top}$, where $(x,y)$ are the coordinates of the point $P_0$ (see Figure~\ref{fig2}), $\phi$ is the heading angle of the robot, and $\theta_r, \theta_l$ are the angles of the right and the left driving wheels, respectively. If we let $S(q)$ be a full rank matrix formed by a set of smooth and linearly independent vectors such that $S^{\top}(q)A^{\top}(q) = 0$ then it is easy to verify that $S(q)$ is given by

\begin{equation}
S(q) = 
\begin{bmatrix}
\frac{r}{2b}(bcos(\phi)-dsin(\phi)) & \frac{r}{2b}(bcos(\phi)+dsin(\phi)) \\
\frac{r}{2b}(bsin(\phi)+dcos(\phi)) & \frac{r}{2b}(bsin(\phi)-dcos(\phi))\\
\frac{r}{2b}  & -\frac{r}{2b} \\
1 & 0 \\
0 & 1
\end{bmatrix}
\label{transformation}
\end{equation}

Then according to Equation~\ref{dynamics} and the fact that $S^{\top}(q)A^{\top}(q) = 0$, it is straightforward to find that 

\begin{equation}
\dot{q} = S(q)\omega
\label{kinematics}
\end{equation}
where $\omega = [\omega_r \omega_l]$ is the vector of angular velocities of the right and left wheel, respectively. Equation~\ref{kinematics} is the kinematic model of the robot. For the interested reader, differentiating Equation~\ref{kinematics} and substituting the result in Equation~\ref{dynamics}, and then multiplying by $S^{\top}$ we can, of course, eliminate the constraint matrix $A^{\top}(q)\lambda$ and obtain the dynamic model of the robot in the form 

\begin{equation}
\bar{M}(q)\dot{\omega} + \bar{V_m}(q, \dot{q})\omega = \bar{B}(q)\tau
\label{controldynamics}
\end{equation}
where $\bar{M} = S^{\top}MS$, $\bar{V_m} = S^{\top}(M\dot{S} + V_mS)$ and 
\begin{equation}
\bar{M}(q) = 
\begin{bmatrix}
\frac{r^2}{4b^2}(mb^2+I) + I_w & \frac{r^2}{4b^2}(mb^2-I) \\
\frac{r^2}{4b^2}(mb^2-I) & \frac{r^2}{4b^2}(mb^2+I) + I_w
\end{bmatrix},
\label{inertia}
\end{equation}
\begin{equation}
\bar{V_m}(q) = 
\begin{bmatrix}
0 & \frac{r^2}{2b}m_c d \dot{\phi} \\
-\frac{r^2}{2b}m_c d \dot{\phi} & 0
\end{bmatrix},
\label{velocity}
\end{equation}
\begin{equation}
\bar{B} = 
\begin{bmatrix}
1 & 0\\
0 & 1
\end{bmatrix},
\label{gain}
\end{equation}and\\
\begin{equation}
\tau = 
\begin{bmatrix}
\tau_r\\
\tau_l
\end{bmatrix},
\label{tau}
\end{equation}
where $m_c$ is the mass of the robot's body and $m_w$ is the mass of a driving wheel plus its associated motor, $I, I_w$ are the moments of inertia of the body around the vertical axis through $P_c$ and the driving wheel (with a motor) about the wheel axis, respectively.

When considering the dynamic model in Equation~\ref{controldynamics}, accurate knowledge about the parameters values of the mobile robot dynamics is nearly impossible to obtain in practice. If we consider that these parameters are also time varying, the problem becomes even more complicated. It was originally proven in the work of \cite{bloch2015introduction} that a continuous (smooth) time-invariant pure state feedback rule, resulting from a violation of Brocketts' necessary condition for stability, cannot stabilize a nonholonomic system to a single equilibrium point. Furthermore, a wheeled mobile robot is only locally controllable over short time intervals, according to \cite{bloch2015introduction}, and it is a controllable system independent of the nature of the nonholonomic constraints $A^{\top}(q)\lambda$ \, as shown by \cite{campion1991modelling}. As a result, the control options are either (a) discontinuous time invariant feedback laws or (b) continuous but time variable non linear feedback control laws applied on the model in Equation~\ref{kinematics}. More precisely, for the controller design we will use the explicit form of Equation~\ref{kinematics}

\begin{equation}
\frac{d}{dt}
\begin{bmatrix}
x\\
y\\
\phi\\
\theta_r\\
\theta_l
\end{bmatrix}=
\begin{bmatrix}
\frac{r}{2}\cos(\phi) & \frac{r}{2}\cos(\phi)\\
\frac{r}{2}\sin(\phi) & \frac{r}{2}\sin(\phi)\\
\frac{r}{2b} & -\frac{r}{2b} \\
1 & 0 \\
0 & 1
\end{bmatrix}
\begin{bmatrix}
\omega_r\\
\omega_l
\end{bmatrix}
\label{controlmodel}
\end{equation}
and given the known relation between the linear $v$ and angular $\omega$ velocities of the robot and the individual wheel angular velocities $\omega_r, \omega_l$ (i.e. knowing the wheel radius and distance between wheels), we can rewrite Equation~\ref{controlmodel} as the ordinary form of a mobile robot with two actuated wheels in 

\begin{equation}
\frac{d}{dt}
\begin{bmatrix}
x\\
y\\
\phi
\end{bmatrix}=
\begin{bmatrix}
\cos(\phi) & 0\\
\sin(\phi) & 0\\
0 & 1
\end{bmatrix}
\begin{bmatrix}
v\\
\omega
\end{bmatrix}
\label{ordinarycontrolmodel}
\end{equation}

Now, with all the modeling in place, we reiterate the objective of trajectory tracking as a control synthesis problem to compute the velocity of the robot such that its pose $P_r = [x_r,y_r,\phi_r]^{\top}$ follows a reference trajectory of the virtual robot $P_d = [x_d,y_d,\phi_d]^{\top}$.

The problem of trajectory tracking for fully actuated systems is now well known, and adequate solutions may be found in advanced nonlinear control textbooks. However, in the case of underactuated vehicles, that is, vehicles with fewer actuators than state variables to be tracked, the problem is still a hotly debated research topic. Linearization and feedback linearization algorithms from \cite{godhavn1997lyapunov} and \cite{walsh1994stabilization} have been developed, as have Lyapunov-based control laws, with representative designs in the work of \cite{wit1993nonlinear} and \cite{fierro1997control}. Independent of the synthesized control law, the trajectory tracking problem can be graphically formulated as shown in Figure~\ref{fig4}. 

Now, putting all elements together, we assume that a feasible desired trajectory for the mobile robot is pre-specified
by a velocity planner from \cite{solea2007trajectory} and fed to a closed-loop control system that will ensure that the robot
will correctly track the desired trajectory under a large class of disturbances. The motion of the robot, following the models above and the conventions in Figure~\ref{fig4}, is given by Equations~\ref{controlmodel} and ~\ref{ordinarycontrolmodel}.

\begin{equation}
\left\{
\begin{array}{ll}
\dot{x}_r(t) = v_r(t)\cos(\phi_r(t)) \\
\dot{y}_r(t) = v_r(t)\sin(\phi_r(t)) \\
\dot{\phi}_r(t) = \omega_r
\end{array}
\right.
\label{controlmodelmotion}
\end{equation} 
where $x_r$ and $y_r$ are the Cartesian coordinates of the geometric center of the mobile robot, $v_r$ is the linear
velocity of the robot, $\phi_r$ is the robot's heading angle, and $\omega_r$ is the angular velocity of the robot, respectively. The trajectory tracking errors can be described by the vector $(x_e, y_e, \phi_e)$ depicted in Figure~\ref{fig4}. The designed controller needs to generate a command vector $(v_c, \omega_c)$. Considering the ordinary form of the mobile robot in Equation~\ref{controlmodelmotion} the error vector, following the convention in Figure~\ref{fig4}, is given by Equation~\ref{errorvector}.

\begin{equation}
\begin{bmatrix}
x_e\\
y_e\\
\phi_e
\end{bmatrix}=
\begin{bmatrix}
\cos(\phi_d) & \sin(\phi_d) & 0\\
-\sin(\phi_d) & \cos(\phi_d) & 0\\
0 & 0 & 1
\end{bmatrix}
\begin{bmatrix}
x_r - x_d\\
y_r - y_d \\
\phi_r - \phi_d
\end{bmatrix}
\label{errorvector}
\end{equation}
where the vector $[x_d,y_d,\phi_d]^{\top}$ is the virtual robot pose. The corresponding error derivatives are then given by Equation~\ref{derrorvector}.

\begin{equation}
\left\{
\begin{array}{ll}
\dot{x}_e(t) = -v_d + v_r\cos(\phi_e) + y_e\omega_d \\
\dot{y}_e(t) = v_r\sin(\phi_e) - x_e\omega_d\\
\dot{\phi}_e(t) = \omega_r - \omega_d
\end{array}
\right.
\label{derrorvector}
\end{equation} 
where $v_d$ and $\omega_d$ are the desired robot linear and angular velocities, respectively.

A final important component in the robot trajectory tracking control loop is the trajectory planner (i.e. generating the desired trajectory in Figure~\ref{fig3}). Although mobile robots' motion planning has been extensively investigated in recent decades, the need of developing trajectories with minimal related accelerations and jerks is not clearly traceable in the technical literature. In our experiments, we used the excellent work of \cite{1706786} to tackle velocity planning and provide suitable time sequences for use in interpolating curve planners. Using this approach allowed us to develop speed profiles (i.e. both linear and angular) that lead to trajectories that are comfortable for humans, as validated in the study of \cite{solea2007trajectory}.

\subsection{Antifragile control}

\begin{figure}[t]
\includegraphics[scale=0.25]{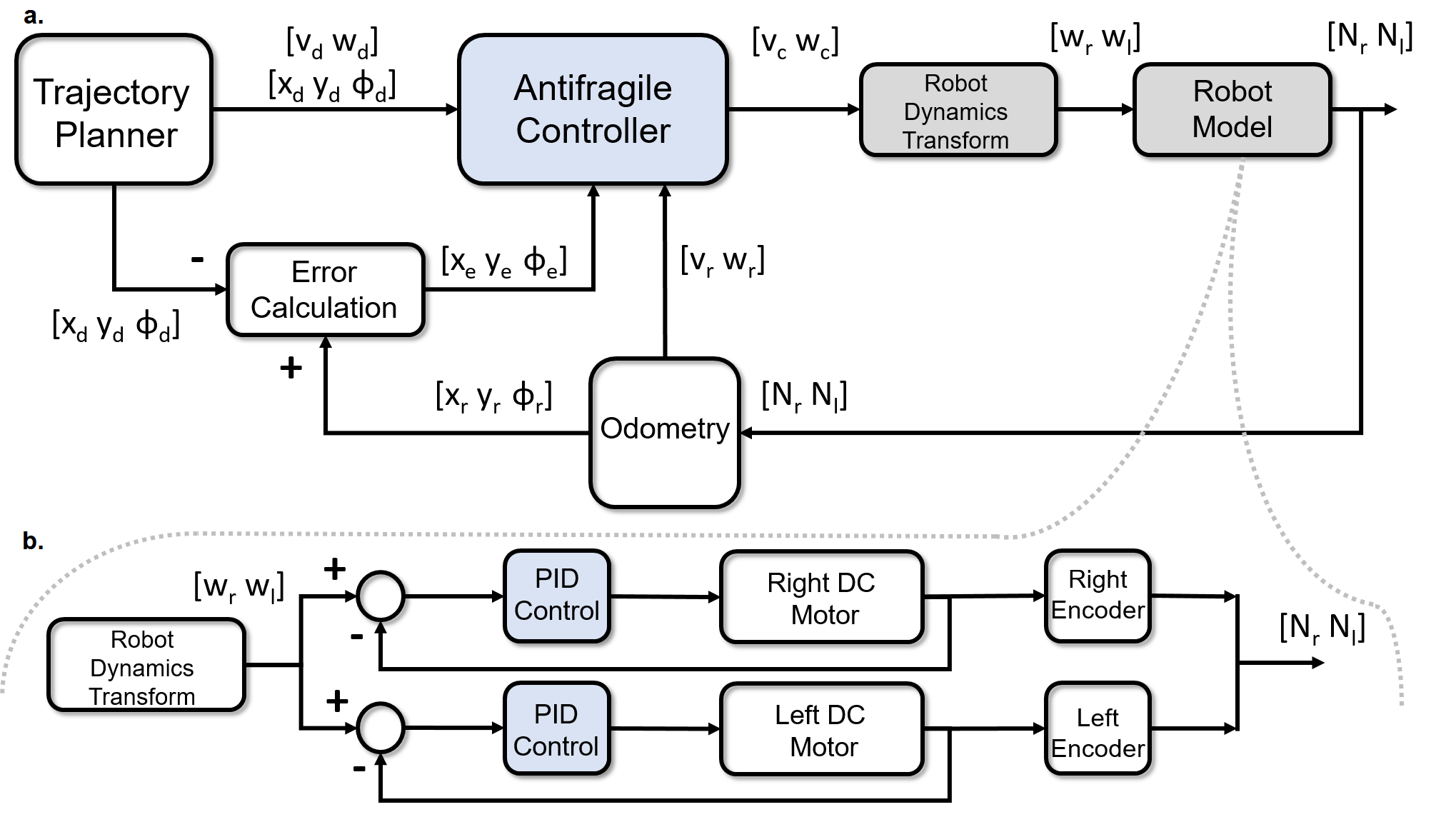}
\centering
\caption{Global control structure for antifragile control of mobile robot trajectory tracking. a) The outer loop contains slower dynamics of pose correction based on the antifragile controller that uses the measured and reference pose vector to compute new linear and angular velocity values. b) The inner loop contains faster dynamics of the two actuators (i.e. DC motors) which control the individual angular velocities of the wheels based on the motion of the motors measured through position encoders. Notations and conventions are consistent with Figure~\ref{fig4} and the equations in the section. The Robot Dynamics Transform is the inverse kinematics transformation from $[v_c, w_c]$ to $[w_r, w_l]$.}
\label{fig5}
\end{figure}

This section is dedicated to introducing the mathematical apparatus of antifragile control, going from its theory and principles to the control synthesis for robot trajectory tracking under uncertainty. The control loop in Figure~\ref{fig3} is effectively expanded in Figure~\ref{fig5} in order to introduce the synthesis of the antifragile controller, based on Equation~\ref{controldynamics}, Equation~\ref{controlmodel}, and Equation~\ref{transformation}, respectively. 

\subsubsection{Preliminaries}

In this subsection, we provide a brief overview of the principles, theory, and design of antifragile control systems, with a particular emphasis on the robot trajectory tracking problem. As defined in Taleb's book \cite{taleb2012antifragile}, antifragility is a feature of a system that allows it to benefit from uncertainty, unpredictability, and volatility, in contrast to fragility. The reaction of an antifragile system to external perturbations is beyond robust and resilient such that mild stresses or perturbations can increase the system's future response by adding a significant anticipation component.

When it comes to control systems, producing such behavior (i.e., induced antifragility) in a feedback control loop provides for a unique design and synthesis method in which: 1) \textit{redundant overcompensation} may drive the system into an overshooting mode that accumulates extra capacity and capability in anticipation; 2) \textit{structure-variability} can elicit stressors that carry intrinsic information which emerges only under volatility and unpredictability of the system dynamics affected by the application of a high-frequency component; 3) \textit{time scales separation} of the interacting system's dynamics that undergo an order-reduction while driven towards the desired antifragile operation region. We will now explain how the preceding notions relate to the robot trajectory control problem and how they might be applied practically.

\subsubsection{Control synthesis}

Previously, in \cite{axenie2022antifragile} and \cite{axenie2022antifragiletraffic}, we cast the control design in geometric control and Riemannian geometry objects, as formally explained in \cite{lee2006riemannian}. This allowed us to work in a coordinate-free environment, relying on the embedding of a manifold into a wider dynamical space, which allowed for simpler control law definitions adequate for manifolds with curvature. In this study, we consolidate this approach and reduce several previous assumptions, while giving concrete control synthesis steps. For the interested reader, more theoretical insights on casting the antifragile control theory in the Riemannian geometry framework are introduced in \cite{axenie2022antifragile}.

We aim at designing a controller that forces the robot to track a prescribed trajectory (i.e., a velocity-parametrized reference temporal evolution) with certain geometrical properties. The problem can be also formulated to compute a control signal (i.e., reference linear and angular velocities) such that the robot dynamics state trajectory confines itself to a desired dynamics where the error vector $(x_e, y_e, \phi_e)$ is minimized. In other words, we want to drive the closed-loop system state evolution to a manifold such that the longitudinal $x_e$, the lateral error $y_e$, and the angular error $\phi_e$ are internally mutually coupled on the considered manifold leading to convergence of all three variables.

In our control synthesis, we decouple the two internal control loops in Figure~\ref{fig5} (see the darkly shaded boxes termed Antifragile Control and PID control) in order to describe the specific design steps of 1) redundant overcompensation; 2) structure-variability; and 3) time scales separation for uncertainty isolation. 

\subsubsection{Time scale separation}

Given the interactions between the two nested control loops (see Figure~\ref{fig5} internal DC motor control loop and the outer position control loop), in order to handle uncertainty and high-frequency phenomena, we need to enable the closed-loop system to separate the time scales of the loops. A very useful tool for such interventions in closed-loop control is (singular) perturbation theory, initially proposed by \cite{fenichel1979geometric} and further extended in \cite{jones1995geometric}. Within this framework, the high-frequency dynamics are taken into account by considering them in a separate timescale. This transformation is achieved by a dynamic change in the order of the controlled system  as a parameter perturbation (akin to a parallel transport map on the Riemannian manifold of the system state trajectory).

Such a change in the controlled system dynamics is more "abrupt" than a normal perturbation to which the system is exposed -- hence the singular perturbation. The main argument in using such an approach in our antifragile control design resides in the fact that such "parasitic", high-frequency phenomena are able to build capacity in reacting to high-amplitude changes in the robot's operation (e.g., wheel slipping, flat tire, shaft bending). The goal of this section is to introduce the reader to how time scale separation through singular perturbation theory is a component of antifragile control and how can be practically used for controller synthesis.

\begin{figure}[H]
\includegraphics[scale=0.4]{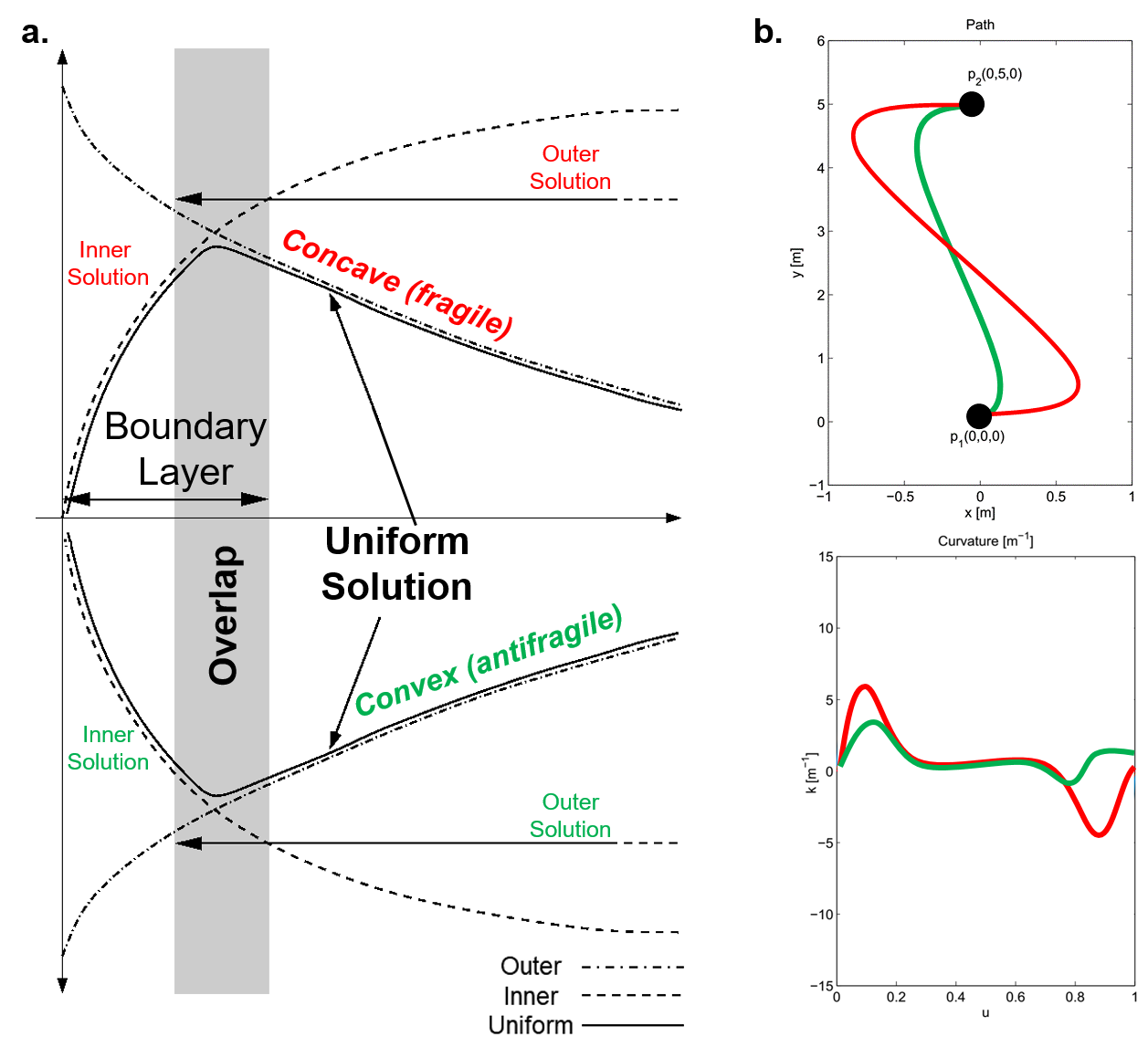}
\centering
\caption{Singular perturbation for time scale separation in antifragile controller synthesis. Using boundary layers and matched asymptotic expansions to probe antifragile behavior. a) Generic depiction of a boundary layer and the types of solutions in singularly perturbed dynamic systems. b) Mapping the boundary layers and shape (convexity/concavity) of the solution to robot velocity planning akin to the desired dynamics to track in the presence of uncertainty.}
\label{fig6}
\end{figure}

The core idea of time scale separation is to capture the dominant phenomena dynamics and then capture the stressors and is typically achieved by "outer" series expansions or  "inner" boundary layer expansions, as suggested in \cite{hunter2004asymptotic} and graphically depicted in Figure~\ref{fig6}. 

More precisely, considering singularly perturbed dynamical systems, we can benefit from solutions with fast variation zones. These areas, which may be seen in the solution or its derivatives, are referred to as "layers", and they frequently exist near the domain border, as shown in Figure~\ref{fig6}. Constructing a solution to a differential equation of a dynamical system entails multiple steps, including determining the locations of layers (whether border or internal), obtaining asymptotic approximations to the solution in different regions (corresponding to distinct differentiated limits in the equations), and finally producing a uniformly valid solution throughout the whole domain, as described in \cite{kokotovic1999singular} and depicted in Figure~\ref{fig6} a. Inner solutions are found for the layers, whereas outer solutions are obtained for the regular distinguished limits. The uniform solution is described by the curvature (i.e., 2-order effect) of the overlapping region between the inner and the outer layer. Interestingly, this can be exploited in our design to define fragile and antifragile control regimes depending on the curvature in the overlap region of the solutions (i.e., attractors/solutions in the system's state space). As depicted in Figure~\ref{fig6} a, we define the antifragile region as the convex region of the solution curve. Hence, the closed-loop system response is antifragile if the curvature is negative, otherwise is fragile (see Figure~\ref{fig6} a).

To be more specific, in our robot trajectory tracking control problem, the reference trajectory is a path, which is an explicit function of time (see Figure~\ref{fig6} b). To achieve a smooth robot movement, the trajectory must be twice differentiable to give a continuous velocity and acceleration. As a result, curve fitting is an integral part of trajectory planning. The most effective method, as demonstrated in \cite{solea2009sliding}, is the use of piece-wise quintic polynomials, also known as quintic splines. These quintic splines are ideal since they provide continuity in position, heading, curvature, velocity, and acceleration. In our experiments, we used the method of \cite{solea2009sliding} to obtain longitudinal and angular velocity profiles ($v(t)$ and $\omega(t)$). The profile must be compatible with the characteristics of the actuators of the robot (i.e., DC motor regimes) and assigned total path length, and it must comply with human comfort travel. 

Let's now intuitively explore the mapping between the shape of the solution of the dynamics of the robot, depicted in Figure~\ref{fig6} a, to the actual control signals $v(t)$ and $\omega(t)$ needed to track the prescribed trajectory, depicted in Figure~\ref{fig6} b upper panel. This will support the need for such a design component in inducing an antifragile behavior. Figure~\ref{fig6} a depicts the solution space of the planned trajectory of the robot given the explicit expression of curvature $k_{i, i+1}$ between consecutive points $i$ and $i+1$ as
\begin{equation}
k_{i, i+1}(t) = \frac{\dot{x}_{i, i+1}\ddot{y}_{i, i+1} - \ddot{x}_{i, i+1}\dot{y}_{i, i+1}}{\dot{x}_{i, i+1}^2+\dot{y}_{i, i+1}^2},
\label{curvature}
\end{equation}
whose solution must be compatible, as mentioned above, with the DC motor regimes, assigned total path length, and it must comply with human comfort travel. Based on the solutions of Equation~\ref{curvature} and their sign (i.e., antifragile control signals for tracking if the curvature is negative, otherwise fragile), the computed signals $v(t)$ and $\omega(t)$ to move the robot from point $p_1$ to point $p_2$ will generate two possible paths, \textit{fragile} (red) and  \textit{antifragile} green (Figure~\ref{fig6}). 
Both fragile (red) and antifragile (green) are feasible solutions. The antifragile solution will limit the curvature and, implicitly, the magnitude of the control input $v(t)$ and $\omega(t)$. This reduces the stress on the robot's actuators and ensures higher robustness in the case of uncertainty while maintaining comfort. The fragile path and curvature will have more prominent curvature variation at the beginning and at the closing of the spline respectively, which might reduce robustness in case of, for instance, wheel slippage or mechanical damage during navigation.

As we have shown in the previous antifragile control instantiation of \cite{axenie2022antifragiletraffic}, oscillators are periodic dynamical systems having "rapid" (inner layers) and "slow" (outer) dynamics (see the principle in Figure~\ref{fig6}). We then anticipated that a uniformly valid solution may be produced by asymptotic matching of the inner and outer solutions, which is based on the fundamental premise that the various solution forms overlap at some recognizable location, typically provided by the curvature (see Figure~\ref{fig6}).

\subsubsection{Redundant overcompensation}

In the following, we revisit the core idea of time scale separation, graphically depicted in Figure~\ref{fig6}. Let us consider a more general form of the robot model as
\begin{equation}
\left\{
\begin{array}{ll}
\dot{x} = f(x, z, \varepsilon, t), ~x(t_0) = x^0, ~x\in \mathbb{R}^n \\
\varepsilon\dot{z} = g(x, z, \varepsilon, t), ~z(t_0) = z^0, ~z \in \mathbb{R}^m
\end{array},
\right.
\label{singpert}
\end{equation} 
where $f, g$ are continuous differentiable functions of $x, z, \varepsilon, t$, basically accounting for the robot model in Equations~\ref{controlmodel}. The scalar $\varepsilon$ quantifies all small parameters of the system (i.e., $I_m$, $I_w$, etc.), which in the antifragile control framework are termed as stressors for capacity build-up. Furthermore, if we consider $T_1$ and $T_2$ two small time constants of the same order of magnitude, we can assume that they can be taken as $\varepsilon$ and have, let's say $T_1 = \varepsilon$ and $T_2 = \alpha \varepsilon$, where $\alpha = T_2/T_1$ is a known constant. Now, if we set $\varepsilon = 0$ in Equations~\ref{singpert}, the dimension of the state space of the system reduces from $n+m$ to $n$ because the second equation degenerates into the transcendental equation $ 0 = g(\hat{x}, \hat{z}, 0, t)$ where $z$ can rapidly converge to a root of the transcendental equation due to its velocity $\dot{z} = g/\varepsilon$, which can be high if $\varepsilon$ is small. 

From a more intuitive perspective, the model in Equations~\ref{singpert} is a reduced-order modeling technique, which allows us to convert the robot's dynamics simplification (reduction) into a parameter perturbation, called "singular". The solutions of the "slow" dynamics $x(t, \varepsilon)$ and the "fast" dynamics $z(t, \varepsilon)$ of the singularly perturbed system in Equations~\ref{singpert} consist of a fast boundary layer and a slow quasi-steady-state, as shown in Figure~\ref{fig6}. From the Riemannian geometry perspective of antifragile control, there exists a manifold $M_\varepsilon$ depending on $\varepsilon$ that can be defined in the space $n+m$ of $x$ and $z$ such that $M_\varepsilon : z = \phi(x, \varepsilon)$. This reduces the dimension of the state space by restricting it to remain on $M_\varepsilon$ manifold. This integrates nicely with the variable structure systems formalism which is also a fundamental design component in induced antifragile control.

\begin{figure}[t]
\includegraphics[scale=0.3]{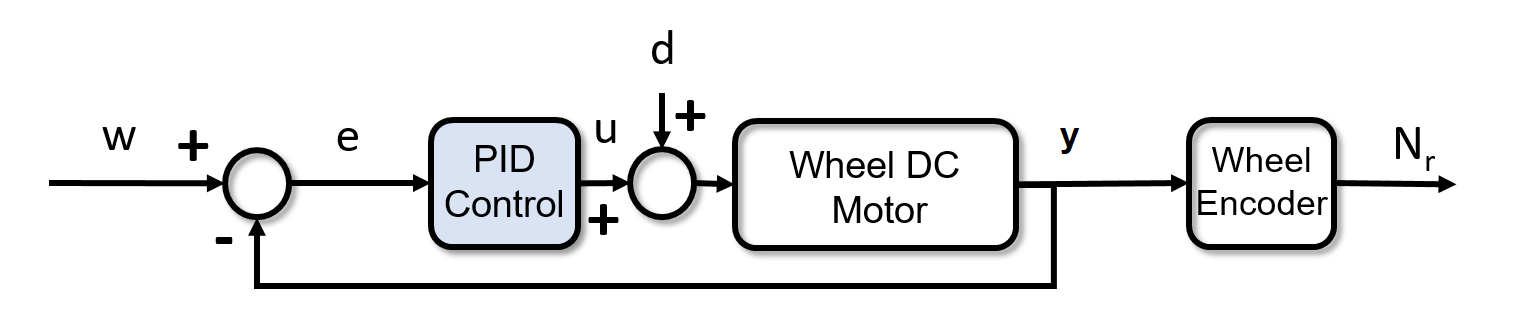}
\centering
\caption{Singular perturbation for low-level inner loop antifragile controller synthesis based on the Proportional Integral Derivative (PID) formulation. Time scale separation can support redundant overcompensation of the low-level actuator control loop. }
\label{fig7}
\end{figure}

Given the formalism introduced in the previous two subsections, we now provide the explicit design and implementation of the time scale separation and redundant overcompensation of the mobile robot antifragile controller. We will consider the inner control loop of the robot actuators in Figure~\ref{fig5} b. Here, we consider the synthesis of the two PID controllers for the closed loop DC motor control, as separately and explicitly shown in Figure~\ref{fig7}. 

We start from the standard formulation of the Proportional Integral Derivative (PID) controllers and a simplified model of the DC motor, as typically found on robotic wheelchairs. In order to focus on the singular perturbation design, we formulate the problem in the Laplace domain of complex frequency $s$. Here, we will only work with algebraic forms of the control law. Later we will come back to the time domain to describe the actual time scale separation.

Referring to Figure~\ref{fig7} and using the typical control theory conventions, the transfer function of the PID controller $u(s)$ is $u(s)/e(s) = K_p(1 + K_Ds + K_I\frac{1}{s})$ where the error $e(s) = w(s) - y(s)$. Furthermore, we consider the DC motor model 
\begin{equation}
\left\{
\begin{array}{ll}
J\dot{\omega} = ki \\
L\dot{i} = -k\omega-Ri+u
\end{array},
\right.
\label{dcmotor}
\end{equation}
where $i, u, R$ and $L$ are the DC motor's armature current, voltage, resistance, and inductance respectively, $J$ is the moment of inertia, $\omega$ is the DC motor shaft angular speed, and $ki$ and $k\omega$ are the torque and the back e.m.f. developed with constant excitation flux. In almost all well-designed motors, $L$ is small and may serve as our parameter $\varepsilon$. This maps to our generic formulation in Equation~\ref{singpert} where $\omega = x$, $i = z$ and the model in Equation~\ref{dcmotor} has the conventional form in Equation~\ref{singpert} when $R \neq 0$. We address the model reduction problem by ignoring the inductance $L$ and solving $ -k\omega - Ri + u = 0$ to obtain the value of the current $i = \frac{u - k\omega}{R}$ which we then substitute in Equation~\ref{dcmotor} in order to obtain the first-order model of the DC motor in the form $J\dot{\omega} = -\frac{k^2}{R}\omega + \frac{k}{R}u$. This basically accounts for finding the manifold $M_\varepsilon$ and restricting the DC motor dynamics to remain on it.

In our design, we need to express that the integral effect of the PID control is much slower than the proportional and the derivative components. The singular perturbation theory supports us in the task of rewriting the control law $u$ given the fact that $K_P$ and $K_D$ offer speed and stability of the system, respectively, whereas $K_I$ reduces the error $e$ to zero. Assuming that $K_I$ is the same order of magnitude with $\varepsilon$, that is, $K_I = \varepsilon\hat{K_I}$, and changing notation as to $k_1 = K_P$, $k_2 = K_PK_D$, and $k_3 = K_P\hat{K_I}$, the control law is $u(s) = (k_1 + k_2s + \varepsilon k_3 \frac{1}{s})e(s)$.

The state variables in the "fast" timescale $\tau = \frac{t}{\varepsilon}$ are $e_1 = e$, $e_2 = \frac{de}{d\tau}$, $e_3=\varepsilon\int_{0}^{\tau}ed\sigma$. Therefore, the state representation of the low-level actuator control loop is described by
\begin{equation}
\left\{
\begin{array}{lll}
\frac{de_1}{d\tau} = e_2 \\
\frac{de_2}{d\tau} = -k_1e_1 - k_2e_2 - k_3e_3 - d \\
\frac{de_3}{d\tau} = \varepsilon e_1
\end{array}.
\right.
\label{dcmotorsingpert}
\end{equation}

If we rewrite the system in the "slow" time variable $t = \varepsilon\tau$ and we identify, following the generic formulation in Equation~\ref{singpert}, selecting $e_1 = z_1$, $e_2 = z_2$ as the fast variables, and $e_3 = x$, as the slow variable respectively, then we have the following formulation of the closed-loop system

\begin{equation}
\left\{
\begin{array}{lll}
\dot{x} = z_1\\
\epsilon\dot{z_1} = z_2 \\
\epsilon\dot{z_2} = -k_1e_1 - k_2e_2 - k_3e_3 - d
\end{array}
\right.
\label{dcmotorsingpertfinal}
\end{equation}
 The fast variables will build capacity for the antifragile response to uncertain events in the robot's operation, such as wheel slipping, flat tire, and DC motor actuator shaft bending. More precisely, we must choose $k_1$ and $k_2$ such that in the PD part of the motor controller the system matrix of Equation~\ref{dcmotorsingpertfinal} is Hurwitzian, that is 
\begin{equation}
\mathrm{Re}\left\{ \lambda\left(
\begin{bmatrix}
0 & 1\\
-k_1 & -k_2
\end{bmatrix}\right)\right\} < 0~.
\label{hurwitz}
\end{equation}

Finally, the integral (I) component of the reduced order (dominant) PID controller is obtained by setting the fast variables to $z_1 = -\frac{k_3e_3 + d}{k_1}$ and $z_2 = 0$ (recall that $e_1 = z_1$ and $e_2 = z_2$) such that the state evolution is $\dot{x} = -\frac{k_3}{k_1}x - \frac{d}{k_1}$, given the previous notation $\omega = x$, $i = z$. Then, the boundary layer system (the overlap region in Figure~\ref{fig6} where the uniform solution's convexity can be probed) is given by a simplified dynamics as follows

\begin{equation}
\left\{
\begin{array}{ll}
\frac{d\hat{z_1}}{d\tau} = \hat{z_2} \\
\frac{d\hat{z_2}}{d\tau} = -k_1(\hat{z_1} - \dot{x}) - k_2\hat{z_2} - k_3x - d = -k_1\hat{z_1} - k_2\hat{z_2}
\end{array}.
\right.
\label{innerctrlsingpertctrl}
\end{equation}

\subsubsection{Structure variability}

Very similar to the effect of time scale separation and coupled with the low-level redundant overcompensation using singular perturbation theory in the previous two subsections, we now introduce another component of the antifragile control synthesis, namely structure variability. Already known in the community, what makes a hard challenge in control systems design is operating under heavy uncertainty conditions. Variable structure control (VSC) systems offer a very powerful tool for handling uncertainty in closed-loop, as shown in the seminal work of \cite{decarlo1988variable}.

Typically, withstanding uncertainty can be done through "brute force", but, as we know, any strictly enforced equality removes one "uncertainty dimension". So there is always a price to pay for precisely attaining the control goal, as formally described in \cite{slotine1991applied}. VSC offers a suitable tool to handle such controller design, by providing a powerful reaction to minimal deviations from a chosen constraint. Typically, VSC is practically implemented through sliding mode control, introduced by \cite{utkin1977variable} and further extended in \cite{utkin2008sliding}.

Sliding mode controllers ensure that the maximum deviation from a constraint is proportional to the time interval between the system's observations and its design follows a model reduction principle as the singular perturbation theory. Hence, we devise antifragile control with a unified framework to implement redundant overcompensation, time scale separation, and variable structure control.

\begin{figure}[h!]
\includegraphics[scale=0.38]{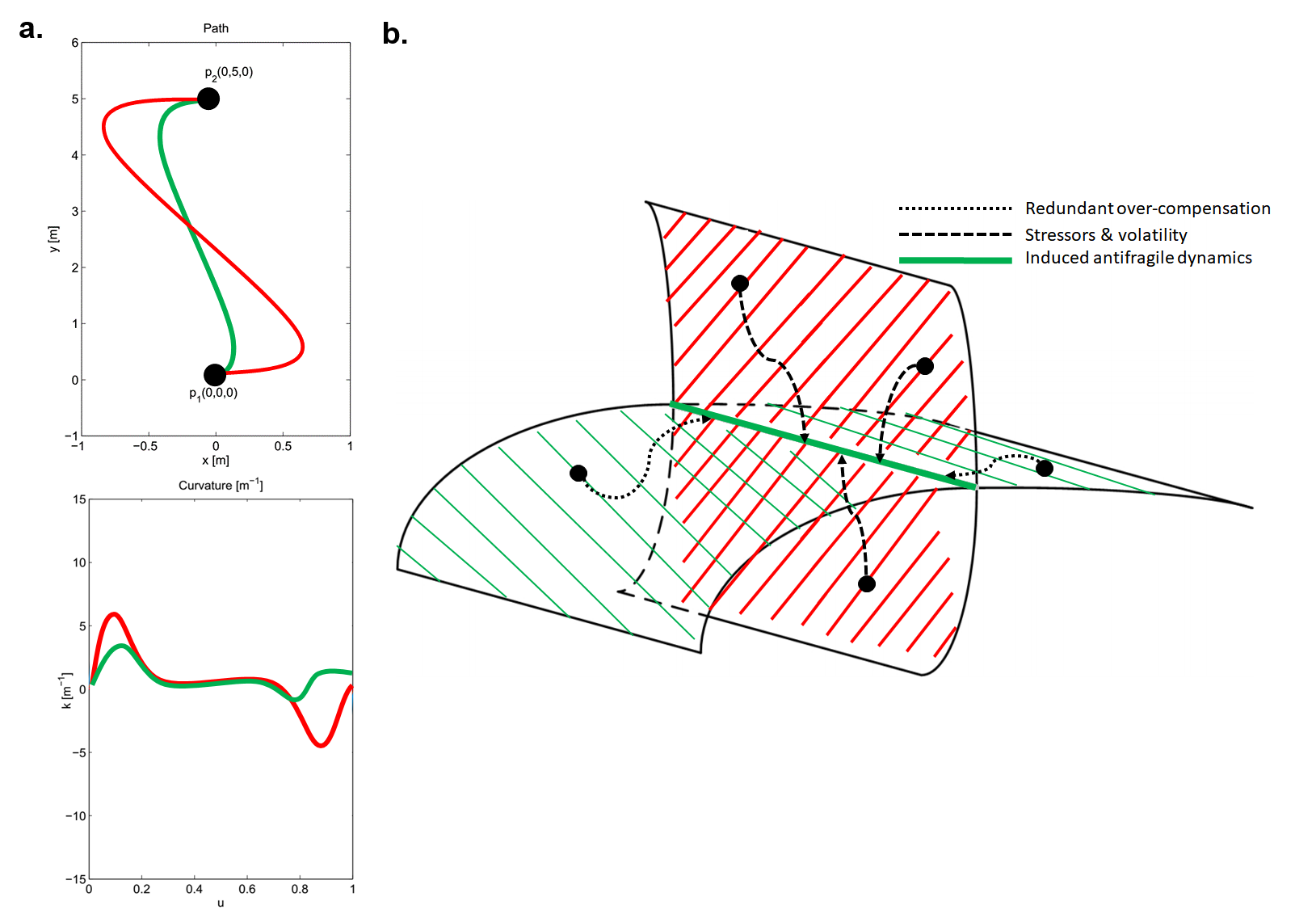}
\centering
\caption{Variable structure control through sliding modes in antifragile control synthesis. a) Robot trajectory tracking control problem: tracking a prescribed trajectory under time constraints and under the effect of uncertainty with comfort constraints (minimal curvature). The fragile (red) dynamics provide feasible dynamics in the trajectory tracking with a control signal which decreases the comfort. The antifragile control signal (green) reaches increased comfort through a feasible and higher-performance (lower error) trajectory tracking performance. b) Closed-loop system dynamics for fragile vs. antifragile behaviors. The induced antifragile control manages to drive the system's dynamics towards the antifragile region (green thick intersecting line of the red and green manifolds) benefiting from redundant over-compensation, stressors, and volatility.}
\label{fig8}
\end{figure}

The advantages of using VSC and sliding mode in our antifragile control design for robot trajectory tracking are listed below:
\begin{itemize}
\item The motion equation of the sliding mode (i.e., the prescribed dynamics), as framed in \cite{slotine1991applied}, can be designed linear and lower-order, despite the fact that robot dynamics and uncertainties effect are highly nonlinear (see Figure~\ref{fig8} b).
\item The sliding manifold (i.e., both a place and a dynamics of the closed-loop robot control) does not depend on the robot model, but it is determined by (problem dependent) parameters selected by the designer, as suggested in \cite{decarlo1988variable}. In our setting, we can force the desired trajectory of the robot in the antifragile region of the prescribed trajectory planning solution with actuator, environment, and comfort constraints (see Figure~\ref{fig6}).
\item Once the sliding motion occurs (i.e., the system dynamics are on the manifold), the robot motion in trajectory tracking has invariant properties which make the motion independent of certain system parameter variations, uncertainty, and disturbances, as described in \cite{utkin1977variable}. Hence, the system performance can be completely determined by the dynamics of the sliding manifold, as depicted in Figure~\ref{fig8}.
\end{itemize}

In order to guide the reader through the intuition behind using VSC for antifragile control synthesis, we consider the simple graphical depiction in Figure~\ref{fig8}. For the robot trajectory tracking problem, the antifragile controller needs to provide a proper control signal (i.e., a pair of longitudinal velocity $v$ and angular velocity $\omega$, such that the path from origin to destination (see Figure~\ref{fig8} a) is tracked under time constraints, uncertainty about the driving surface, actuator failures, and with increased comfort (i.e., minimal curvature). This is achieved through a proper synthesis of the control law, which builds up through a redundant over-compensation capacity to cope with uncertainty about the driving surface and actuator failures (see initial conditions of the system dynamics converging from the green manifold to the induced antifragile dynamics from Figure~\ref{fig8} b). When starting from a fragile region of the system's solutions (see red manifold from Figure~\ref{fig8} b) the controller handles stressors and volatility (i.e. ,increased curvature of the trajectory and then tracking the desired path) by using the provided "inertia" to converge to the induced antifragile dynamics.

Now, in order to attain such dynamics under induced antifragile control, we cover the relevant design steps of a VSC with respect to our problem. 

\paragraph{Sliding manifold selection}

In this control design step, we need to choose a sliding manifold with a lower order than the system such that the system performance is achieved during the sliding motion. This step is highly dependent on the problem, hence we will motivate the choice for the trajectory tracking problem we consider. 
We start from the canonical form of a sliding manifold $s$ depending on system state dynamics $x$ (see Equation~\ref{singpert} for the general dynamical system formulation) in Equation~\ref{surfacedesign}.

\begin{equation}
\dot{s} = \frac{\partial s}{\partial x}\dot{x} = \frac{\partial x}{\partial x}s(x) = \lambda_1x_1 + \lambda_2x_2+ ... +x_n = 0
\label{surfacedesign}
\end{equation}
where the coefficients $\lambda_i$ in $\dot{s}$ define the desired characteristics of the sliding mode, that is the characteristics of the closed-loop system after the manifold reaching phase, as broadly described in \cite{utkin1977variable}. 
Finding these parameters is typically formulated as an optimization problem, and solved using linear programming techniques (e.g., Linear Quadratic (LQ) approach), as shown in \cite{utkin1978methods}. Here, a criteria for a second order system $J = \int_{t_s}^{\infty}(x_1^{\top}Q_{11}x_2 + 2x_1^{\top}Q_{12}x_2 + x_2^{\top}Q_{2}x_2)dt$ was minimized to get the optimal sliding manifold. Considering $Q_{12} = 0$ then the optimal control $x_2 = - Q_{22}^{-1}A_{12}^{\top}Px_1 = -kx_1$ where P is a p.d. matrix solution of the Ricatti equation $A_{11}^{\top}P + PA_{11} - PA_{12}Q_{22}^{-1}A_{12}^{\top}P = -Q_{11}$ where $A$ is the input matrix of the system. The switching function is obtained by simply considering $s(x) = kx_1 + x_2 = [Q_{22}^{-1}A_{12}^{\top}P, I]x$.

In our case, we want to choose a sliding manifold such that the longitudinal error $x_e$, the lateral error $y_e$, and the angular error $\phi_e$ are internally coupled to ensure mutual convergence. Given the robot error in the outer loop (see Figure~\ref{fig4}) 
\begin{equation}
\begin{bmatrix}
x_e\\
y_e\\
\phi_e
\end{bmatrix}=
\begin{bmatrix}
\cos(\phi_d) & \sin(\phi_d) & 0\\
-\sin(\phi_d) & \cos(\phi_d) & 0\\
0 & 0 & 1
\end{bmatrix}
\begin{bmatrix}
x_r - x_d\\
y_r - y_d \\
\phi_r - \phi_d
\end{bmatrix},
\label{errorvectorsurface}
\end{equation}
where the vector $[x_d,y_d,\phi_d]^{\top}$ is the virtual robot pose. The corresponding error derivatives are then given by 
\begin{equation}
\left\{
\begin{array}{ll}
\dot{x}_e(t) = -v_d + v_r\cos(\phi_e) + y_e\omega_d \\
\dot{y}_e(t) = v_r\sin(\phi_e) - x_e\omega_d\\
\dot{\phi}_e(t) = \omega_r - \omega_d
\end{array}.
\right.
\label{derrorvectorsurface}
\end{equation} 
The sliding manifolds we choose for the robot trajectory tracking are 
\begin{equation}
\left\{
\begin{array}{ll}
s_1 = \dot{x}_e + \lambda_1x_e\\
s_2 = \dot{y}_e + \lambda_2y_e + \lambda_0 \mathrm{sgn}(y_e)\phi_e
\end{array}
\right.
\label{trackingsurfaces}
\end{equation} 
with $\lambda_0, \lambda_1, \lambda_2 > 0$.
Interestingly, if $s_1$ converges to 0 then $x_e$ converges to 0. Additionally, if $s_2$ converges to 0, then at steady state $\dot{y}_e = -\lambda_2y_e - \lambda_0\mathrm{sgn}(y_e)\phi_e$. Here, for negative lateral error $y_e < 0$ then $\dot{y}_e > 0$ if and only if $\lambda_0 < \lambda_2\frac{|y_e|}{|\phi_e|}$ and for a positive lateral error $y_e > 0$ then $\dot{y}_e < 0$ if and only if $\lambda_0 < \lambda_2\frac{|y_e|}{|\phi_e|}$.

\paragraph{Control law design}
In this step, we need to design a switched feedback control law that satisfies the reaching condition (see Figure~\ref{fig8} b) and drives the system trajectory to the manifold in finite time and keeps it there thereafter. In this study, we consider Gao's reaching law introduced in \cite{gao1993variable} that employs the differential equation $\dot{s} = -Q\mathrm{sgn}(s) - Ph(s)$, where $Q = \mathrm{diag}[q_1, q_2, \ldots, q_n]$ with $q_i > 0, i = 1, \ldots, n$; $P = \mathrm{diag}[p_1, p_2, \ldots, p_n]$, with $ p_i > 0, i = 1, \ldots, n$; $\mathrm{sgn}(s) = [\mathrm{sgn}(s_1), \mathrm{sgn}(s_2),\ldots, \mathrm{sgn}(s_m)]^{\top}$; $h(s) = [h_1(s_1), h_2(s_2), \ldots, h_m(s_m)]^{\top}$; and $s_ih_i(s)>0$ with $h_i(0) = 0$. The reaching time for $x$ to move from an initial state to the switching manifold $s_i$ is finite and given by

\begin{equation}
T_i = \frac{1}{p_i}\mathrm{ln}\frac{p_i|s_i|+q_i}{q_i}              
\label{reachingtime}
\end{equation}

Now, having the reaching law equation we can determine the control law $u$ that drives the robot on the prescribed trajectory for tracking. In our case, the control law is obtained by computing the time derivative (i.e., the velocity) of $s(x)$ along the reaching mode trajectory (see Figure~\ref{fig8} b) as $\dot{s} = \frac{\partial s}{\partial x}(A(x) + B(x)u) = -Q\mathrm{sgn}(s) - Ph(s)$ where, in the generic form, $A$ is the state transformation matrix and $B$ is the control input gain matrix. We then have the control law given by $u = -( \frac{\partial s}{\partial x}A(x) + Q\mathrm{sgn}(s)+Ph(s))( \frac{\partial s}{\partial x}B(x))^{\top}$. In this case, the resulting sliding mode is not preassigned but rather follows the natural state trajectory of a first-order switching scheme, as shown in \cite{hung1993variable}. Of course, the switching takes place depending on the location in the state space of the initial state, as shown in Figure~\ref{fig8}.

In our particular case, we choose the control law $u$ as
\begin{equation}
\dot{s} = -Qs - P\mathrm{sgn}(s)
\label{controllawmanifold}
\end{equation}
with $P,Q > 0$. Opposite to the approach of \cite{hung1993variable}, we use the proportional term $-Qs$ instead of the $\mathrm{sgn}(s)$ to force the system's state to approach the switching manifold faster when $\dot{s}$ is large, while the discontinuous (magnitude) component is given by $h(s) = \mathrm{sgn}(s)$ in the second term (i.e.. the constant rate reaching).
Now, given the ordinary form for control of the mobile robot
\begin{equation}
\frac{d}{dt}
\begin{bmatrix}
x\\
y\\
\phi
\end{bmatrix}=
\begin{bmatrix}
\cos(\phi) & 0\\
\sin(\phi) & 0\\
0 & 1
\end{bmatrix}
\begin{bmatrix}
v\\
\omega
\end{bmatrix}
\label{ordinarycontrolmodelfinal}
\end{equation}
and the derivative of the manifold Equations~\ref{trackingsurfaces} as
\begin{equation}
\left\{
\begin{array}{ll}
\dot{s}_1 = \ddot{x}_e + \lambda_1\dot{x}_e\\
\dot{s}_2 = \ddot{y}_e + \lambda_2\dot{y}_e + \lambda_0\mathrm{sgn}(y_e)\dot{\phi}_e
\end{array}
\right.
\label{trackingsurfacesderivs}
\end{equation}
we perform simple mathematical manipulations to obtain the control law $u = [v_c, \omega_c]^{\top}$ where the linear acceleration is
\begin{equation}
\dot{v}_c = \frac{1}{\cos(\phi_e)}(-Q_1s_1 - P_1\mathrm{sgn}(s_1) - \lambda_1\dot{x}_e - \dot{\omega}_dy_e - \omega_d\dot{y}_e + v_r\dot{\phi}_e\sin(\phi_e) + \dot{v}_d),
\label{controllawacc}
\end{equation}
and the angular velocity is
\begin{equation}
\omega_c = \frac{1}{v_e\cos(\phi_e) + \lambda_0\mathrm{sgn}(y_e)}(-Q_2s_2-P_2\mathrm{sgn}(s_2) - \lambda_2\dot{y}_e - \dot{v}_r\sin(\phi_e) + \dot{\omega}_dx_e + \omega_d\dot{x}_e) .
\label{controllawomega}
\end{equation}
Note that the sign function $\mathrm{sgn}(\cdot)$ in the control signals can be replaced in the practical implementation by the saturation function $\mathrm{sat}(\cdot)$ with thresholds to reduce the chattering phenomenon. Now, let us define the Lyapunov function candidate
\begin{equation}
V = \frac{1}{2}s^{\top}s.
\label{lyapunov}
\end{equation}
The time derivative $\dot{V}$ is given by 
\begin{equation}
\dot{V} = s_1\dot{s}_1 + s_2\dot{s}_2 = s_1(-Q_1s_1 - P_1\mathrm{sgn}(s_1))+ s_2(-Q_2s_2 - P_2\mathrm{sgn}(s_2))
\label{lyapunovvelocity}
\end{equation}
or in a shorter form
\begin{equation}
\dot{V} = -s^{\top}Qs - P_1|s_1|-P_2|s_2|
\label{lyapunovvelocityfinal}
\end{equation}
For $\dot{V}$ to be negative semi-definite, we choose $Q_i$ and $P_i$ such that $Q_iP_i \geq 0$. Then, given that $V>0$ and that $\dot{V} \leq 0$, the control law is stable in the Lyapunov sense.
Finally, the single-wheel velocity commands for the mobile robot are practically given by
\begin{equation}
\left\{
\begin{array}{ll}
\Omega_r = \frac{v_c + b\omega_c}{r}\\
\Omega_r = \frac{v_c - b\omega_c}{r}
\end{array},
\right.
\label{wheelscmds}
\end{equation}
where, following the conventions in Figure~\ref{fig2}, $r$ is radius of the driving wheels, $b$ is half the distance between the driving wheels, $v_c$ is the computed control velocity, and $\omega_c$ the computed control angular velocity (see Equation~\ref{controllawacc} and Equation~\ref{controllawomega}). These values are subsequently sent to the inner loop of the closed-loop system in Figure~\ref{fig5}, more precisely to the PID controllers (separately treated in the previous section and in Figure~\ref{fig6}) where the encoder revolutions $N_r$ and $N_l$ are available from odometric computations.

\subsection{Competitive control algorithms}

As robot control is a very fruitful field where various control algorithms were validated, we selected competitive approaches for the trajectory tracking task from the most prominent state-of-the-art approaches. Additionally, we considered Figure~\ref{fig1} perspective on how each of the control approaches would perform in the face of uncertainty in order to cover the whole fragility--robustness--antifragility spectrum. All the competitive controllers' implementations are available from the codebase on \href{https://gitlab.com/akii-microlab/antifragile-robot-control/}{GitHub}\footnote{Codebase available at \href{https://gitlab.com/akii-microlab/antifragile-robot-control/}{https://gitlab.com/akii-microlab/antifragile-robot-control/}}. 

\subsubsection{Robust control}

For the robust control, we have chosen sliding mode control, as a very powerful method for robot trajectory tracking control, because it shares the advantages of variable structure controller design. The specific control synthesis is based on the work of \cite{solea2007trajectory}. In our experiments, we will denote the sliding mode controller as ROBUST.
The proposed controller used also variable structure synthesis based on equivalent control
\begin{equation}
\left\{
\begin{array}{ll}
u_{eq1}(t)= \frac{-D_1(t)}{\alpha(t) \cos(\phi_e(t))}\\
u_{eq2}(t)= \frac{-D_2(t)}{\beta}
\end{array},
\right.
\label{robustctrl}
\end{equation}
where  $r$ is the wheel radius, $\alpha = 1/rm(t)$ and $\beta(t) = b/rI(t)$ uncertainty parameters in mass $m$ and inertia $I$, and $2b$ is the robot's base width. In Equation~\ref{robustctrl}, $D_1$ and $D_2$ are two functions of the kinematic error derivative in Equation~\ref{derrorvector}.

\subsubsection{Adaptive receding horizon control}

In order to approach adaptive receding horizon control, we considered model predictive control (MPC) as a suitable candidate given its prediction capabilities which contrast well with the anticipation capabilities of the antifragile controller. More precisely, we based our design on the controller design of \cite{wang2019path}. The proposed controller not only provides increased tracking accuracy but also takes the robot's dynamic stability into account throughout the tracking process, i.e., the robot dynamic model is employed as the controller model. Furthermore, the problem of driving comfort created by the usage of a traditional MPC controller when the vehicle deviates from the desired course is resolved by adaptively increasing the weight of the cost function. In our experiments, we denote the model predictive controller as ADAPTIVE.
The purpose of MPC-based trajectory tracking control is to ensure that the error between the predicted output variables and the reference
values as small as possible, which means that the robot can follow the target trajectory accurately and obtain lateral stability. Therefore, the cost function was constructed as follows
\begin{equation}
J = \| Q({y(t)} - \hat{y}_{ref}(t))\|^2 + \| R u(t)\|^2
\label{adaptivectrl}
\end{equation}
where $Q$ and $R$ are weighting matrices of the controlled outputs and inputs, $y(t)$ is the 2D location and the heading angle, $\hat{y}_{ref}(t)$ consists of the reference location and the heading angle in prediction horizon, and $u(t)$ is the control input vector.
\subsubsection{Resilient control}
Finally, in order to represent resilient controller design, we consider a fuzzy logic controller which provides an effective approach to approximate any smooth nonlinear dynamics in the form of IF–THEN rules. The concrete implementation we considered in our experiments is based on the work of \cite{antonelli2007fuzzy}. This work presents a trajectory-tracking strategy based on a fuzzy-logic set of rules that mimics human driving behavior. The fuzzy system's input is estimated information about the next curve ahead of the robot; the related output is the cruising velocity that the robot must achieve in order to safely travel on the path in the allocated time. In our experiments, we will denote the fuzzy logic controller as RESILIENT. For the actual implementation, we used a 4-rules Takagi-Sugeno-Kang fuzzy inference system such that the output is already given in a crisp format, directly applicable to the robot's actuators. The actual control signal is
\begin{equation}
u = [v_l; v_r], u(t,i) = [k_d(t)d_e + k_t(t)\theta_e; k_d(t)de - k_t(t)\phi_e],
\label{resilientctrl}
\end{equation}
where $v_l$ and $v_r$ are the left and right velocities, $d_e$ is the Euclidian distance error in Cartesian space, $\phi_e$ is the heading error of the robot, and $k_d$ and $k_t$ positive sub-unit proportional gain factors.

\section{Experiments and Results}

In order to evaluate the control strategies and demonstrate the benefits that an antifragile design brings, we have designed a systematic analysis and evaluation framework. After modeling and theorizing the induced antifragile control synthesis in the previous section, we dedicate the first part of the current section to the faults and uncertainty injection system. This systematic framework:
\begin{itemize}
\item generates reference trajectories for the closed-loop robot control,
\item supports the induction of user-defined uncertainty injection e.g. wheel slippage, actuator fault, or sensor fault,
\item supports the parametrization (i.e. timing, duration, amplitude, frequency) of user-defined uncertainty injection, and
\item compares the performance when different types of uncertainty and/or faults are injected.
\end{itemize} 

All the experiments, analysis, and additional experiments not discussed in this paper, can be reproduced through the codebase available on \href{https://gitlab.com/akii-microlab/antifragile-robot-control/}{GitHub}\footnote{Codebase available at: \href{https://gitlab.com/akii-microlab/antifragile-robot-control/}{https://gitlab.com/akii-microlab/antifragile-robot-control/}}. 

\subsection{Ideal trajectory tracking robot control}

\begin{figure}[t!]
\includegraphics[scale=0.36]{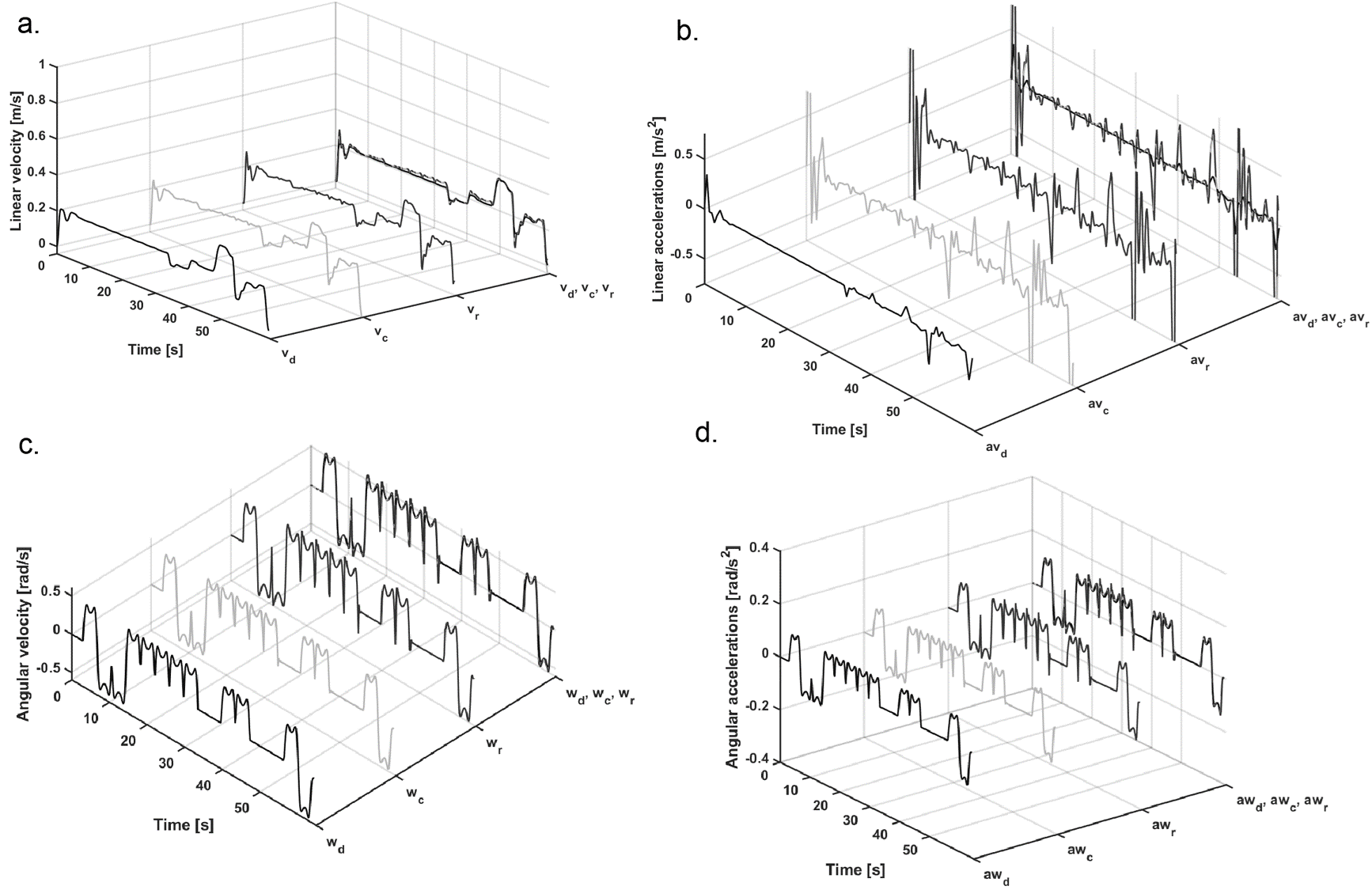}
\centering
\caption{Trajectory tracking ANTIFRAGILE control analysis: a) Linear velocity analysis comparing the desired velocity $v_d$, the computed control signal velocity $v_c$ and the real velocity in closed-loop $v_r$; b) Linear acceleration analysis comparing the desired longitudinal acceleration $av_d$, the computed control signal acceleration $av_c$ and the real acceleration in closed-loop $av_r$; c) Angular velocity analysis comparing the desired angular velocity $w_d$, the computed control signal angular velocity $w_c$ and the real angular velocity in closed-loop $w_r$; d) Angular acceleration analysis comparing the desired angular acceleration $aw_d$, the computed control signal angular acceleration $aw_c$ and the real angular acceleration in closed-loop $aw_r$.}
\label{fig10}
\end{figure}

In the first part of the experiments and evaluation, we delve into the vanilla trajectory tracking robot control results. We analyze here the basic behavior of the selected control approaches on the basic (fault-free) task. We will get some insights into the robot's kinematics and dynamics and the intuition of how each control solves the trajectory tracking. 
Additionally, in order to extract some insights into the actual kinematic and dynamic parameters of the robot under trajectory tracking control, we performed a short analysis of how the closed loop using the ANTIFRAGILE controller. We observed that profiles of the velocities and accelerations are very close to the profile of the reference ones (see Figure~\ref{fig10} a, b, c, d).
\begin{figure}[t!]
\includegraphics[scale=0.32]{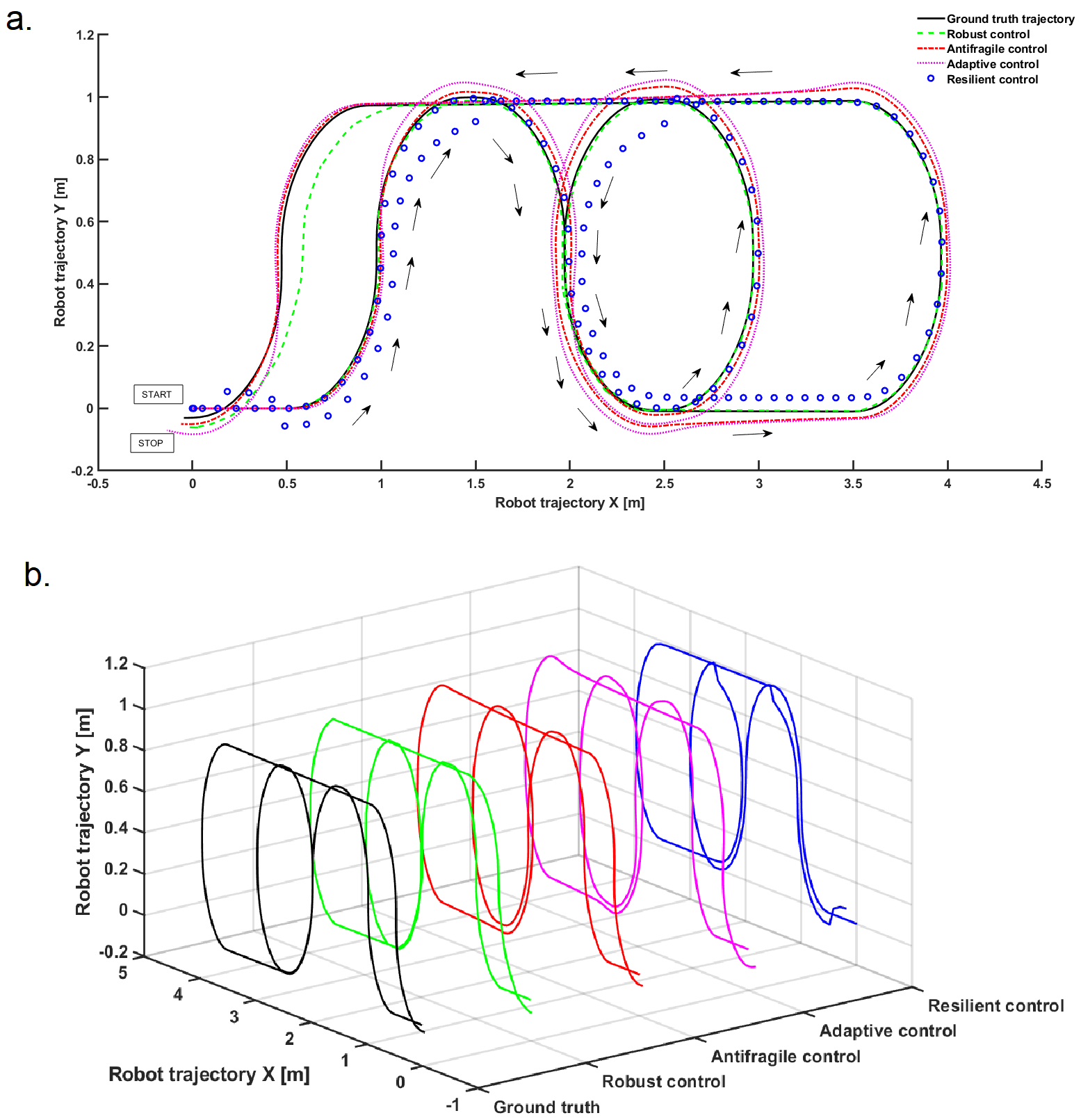}
\centering
\caption{Trajectory tracking control: a) Experimental trajectory description and individual controllers operation. b) Comparison of the control approaches and their performance in tracking the desired trajectory. We can see that each control strategy overshoots when tracking the reference trajectory, but as we will see in the experiments this is a compensation mechanism for the curvature of the trajectory with more capacity to handle uncertainties (i.e., ANTIFRAGILE and ADAPTIVE in panel a)). Additionally, we see undershooting behavior in control approaches which are tracking almost perfectly the low curvature regions (i.e., ROBUST and RESILIENT in panel a) but then have no capacity when the curvature increases (e.g., see end of trajectory and the inner cycle).}
\label{fig9}
\end{figure}
We can already identify in Figure~\ref{fig10} the trademarks of ANTIFRAGILE control in the actual control signals for the robot motion. For instance, the rate of change of linear velocity is determined by the variable structure control component of the antifragile control, which determines capacity building in handling fast-changing curvature values visible through an overshoot in the trajectory (see also Figure~\ref{fig9} a - rightmost loop). Interestingly, the capacity-building feature of the ANTIFRAGILE controller is active (i.e., overshooting) when exiting a section of the trajectory from high curvature to low curvature, whereas the anticipation feature is active when exiting a section of the trajectory from low curvature to high-curvature (see Figure~\ref{fig9} a - leftmost loop toward STOP).
\begin{figure}[t!]
\includegraphics[scale=0.38]{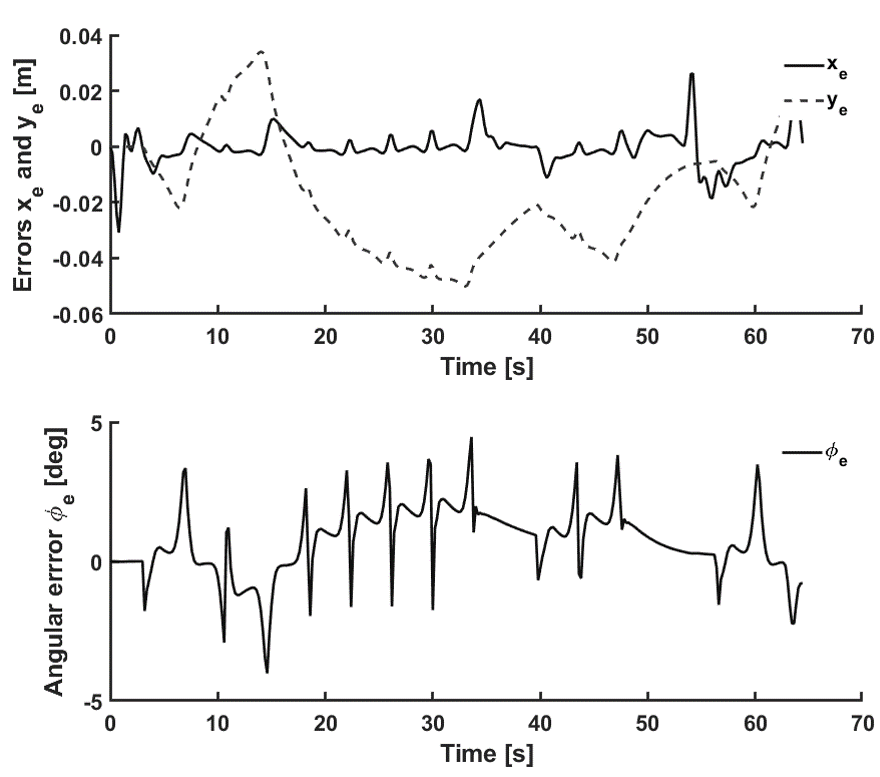}
\centering
\caption{Trajectory tracking ANTIFRAGILE control analysis on the kinematic performance: the longitudinal error $x_e$, the lateral error $y_e$, and the heading error $\phi_e$.}
\label{fig11}
\end{figure}
When considering the kinematic assessment of the ANTIFRAGILE trajectory tracking control in Figure~\ref{fig11}, we can see that the longitudinal error $x_e$ is kept around $0$, with a deviation of maximum 3 cm, whereas the lateral error $y_e$ varies largely due to the often changes in the direction of the robot (see Figure~\ref{fig9}). However, the controller compensates jointly through the variable structure component (i.e. high-frequency changes in the linear and angular velocity control signal in Figure~\ref{fig10}), for both $x_e$ and $y_e$, for an overall 5 cm maximum deviation. The angular error $\phi_e$ is also kept low, with a maximum deviation of 4 degrees, despite the highly curved trajectory in Figure~\ref{fig9}, which is only attained due to the time scale separation and redundant overcompensation components of the ANTIFRAGILE control.

\subsection{Faults and uncertainty injection system}

The fault and uncertainty injection system is based on our previous work in \cite{axenie2010adaptive} and extended in \cite{axenie2010new}. The core idea is to model uncertainty and faults through a series of Kalman filters, basically producing an estimate of the robot state in the presence of uncertainty and faults by encapsulating altered dynamics of the robot according to the type of uncertainty and fault. 
In other words, we exploit the Kalman filter capability to use a set of prediction-correction equations implementing an optimal estimator, by minimizing the estimate error covariance when certain conditions are satisfied. More precisely, in our experiments, each of the Kalman filters encapsulates a kinematic model of the robot but with different parameters (i.e., corresponding to various forms of uncertainty or faults). 

The core idea behind this framework resides in the fact that for the same input vector (with noise), each filter computes a prediction of the robot’s state vector. Each filter is associated with a certain form of uncertainty or fault. We, hence, consider 1) a nominal filter corresponding to the fault-free robot operation; 2) a filter that contains the same robot kinematic model but with modified parameters to emphasize the right tire flat fault (i.e., the right wheel radius has a smaller value progressively), so that its prediction will be the robot
state vector if a right tire flat occurred; 3) a filter modeling and predicting the dynamics of a left tire flat fault; 4) a filter modeling and predicting an actuator shaft bending of the right wheel; and 5) a filter modeling and predicting an actuator shaft bending of the left driving wheel of the robot, respectively. Besides the state estimate each filter generates a measurement vector estimate during the prediction stage, which is used in the correction stage of the filter.

\subsection{Parametrization}
\begin{itemize}
\item ROBUST control: weighted gain control law for equivalent control; reaching mode with separated $x_e$ and $\phi_e$ surfaces in the sliding mode design; simple PID control parametrization;
\item ADAPTIVE control: weighted MPC output to guarantee both tracking accuracy and ride comfort, which can adjust the weights of cost function adaptively based on lateral position error $y_e$ and heading error $\phi_e$;
\item RESILIENT control: 2 inputs ($d_e$ and $|\phi_e|$), 2 outputs ($v_l$, $v_r$) controller; 4 IF-THEN rules; 2 input space membership values; 2 output space membership values; min aggregation for output;
\item ANTIFRAGILE Control: reaching mode with combined $x_e$, $y_e$ errors and separated $\phi_e$ surfaces in the variable structure component design; singular perturbation PID control parametrization.
\end{itemize}

\subsection{Evaluation}

\begin{figure}[t!]
\includegraphics[scale=0.35]{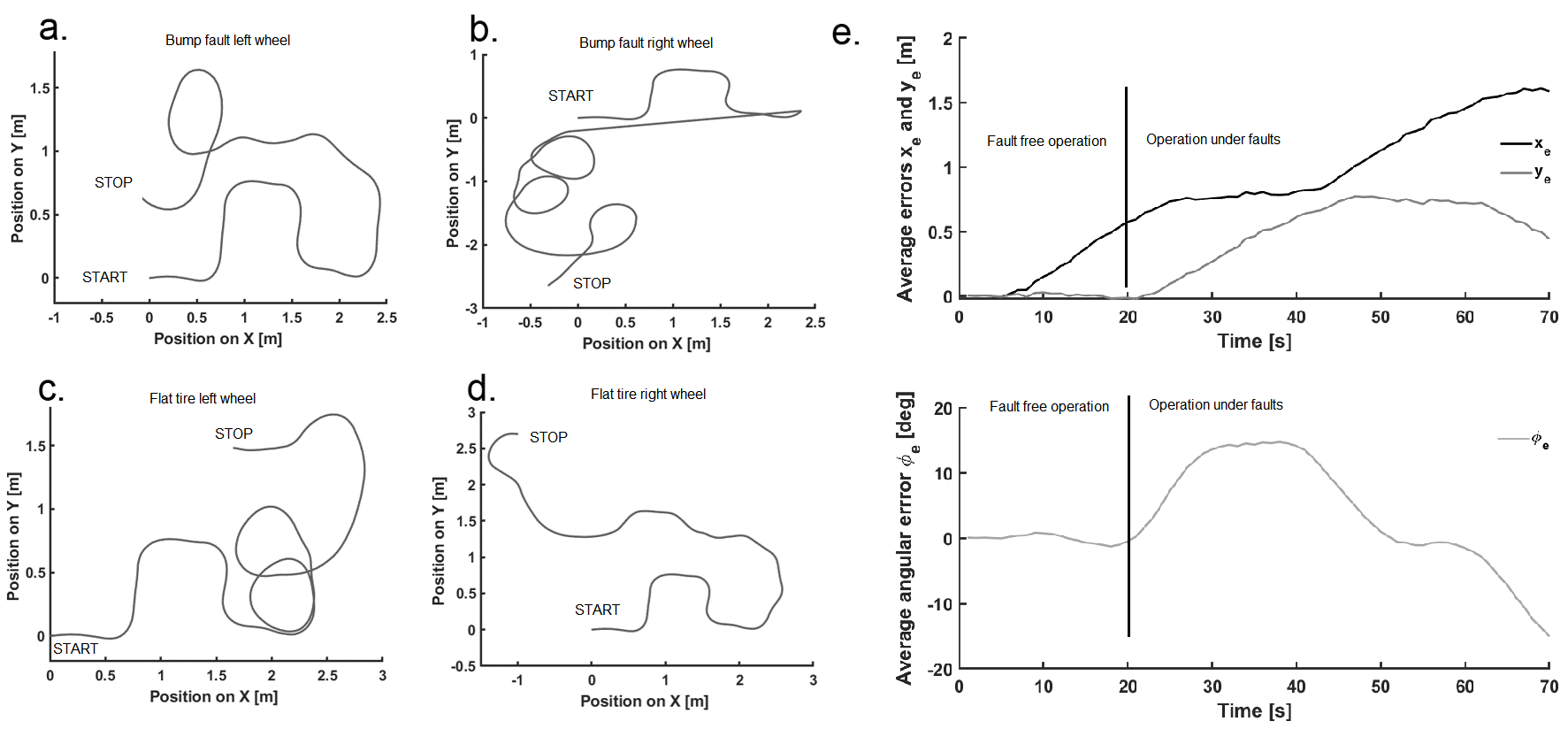}
\centering
\caption{Trajectory tracking analysis in the presence of faults: a) Robot trajectory exposed to a DC motor shaft bump actuator fault on left robot wheel; b) Robot trajectory exposed to a DC motor shaft bump actuator fault on right robot wheel; c) Robot trajectory exposed to a flat tire sensor fault on left robot wheel; d) Robot trajectory exposed to a flat tire sensor fault on right robot wheel; e) The average longitudinal error $x_e$, lateral error $y_e$, and heading error $\phi_e$ over the 4 types of faults. The faults parametrization is as follows: the DC motor shaft bump amplitude is $1.5\,$cm and occurs periodically starting at time $t_{injection} = 20\,$s of trajectory tracking (i.e. after the first half-loop in panels a, b, c, d); the flat tire assumes a time decaying wheel radius decrease from $r_o = 30\,$cm to $r_f = 26\,$cm starting at $t_{injection} = 20\,$s.}
\label{fig12}
\end{figure}

For the evaluation of the four approaches for robot trajectory tracking control, we parametrize the faults and uncertainty injection system, described above, for 4 types of faults/disturbances of the closed-loop system, namely: 2 sensor faults (e.g., perceiving a continuous wheel radius decrease during operation akin to a flat tire) and 2 actuator faults (e.g., a periodic eccentric mechanical motion of the DC motor shaft akin to a wheel bump). The disturbances/faults are amplitude and time parametrized, basically assuming a progressive change in wheel radius over a time period or a fixed amplitude increment of the wheel radius at periodic time intervals. Such parametrization is possible through the Kalman filter bank approach developed by \cite{axenie2010adaptive}. More precisely, the faults parametrization, for the trajectory in Figure~\ref{fig9} and analysis in Figure~\ref{fig12}, is as follows: the DC motor shaft bump amplitude is $1.5\,$cm and occurs periodically starting at time $t_{injection} = 20\,$s of trajectory tracking (i.e., after the first half-loop in panels a, b, c, d); the flat tire assumes a time decaying wheel radius decrease from $r_{fault\_free} = 30\,$cm to $r_{flat\_tire} = 26\,$cm starting at $t_{injection} = 20\,$s.

Intuitively, the effect each type of fault has upon the closed-loop system trajectory tracking control is different and dictated by the effects it has upon the kinematics and the dynamics of the robot. We have analyzed the effects of each of the 4 types of faults, in Figure~\ref{fig12} when considering effects on a simple baseline control scheme (i.e., without any adaptation). This choice is motivated by the fact that we want to understand how each of the faults reflects in the robot's behavior without any means to adapt. 

\begin{table*}\centering
\ra{1.0}
\begin{tabular}{@{}rrrrrrrrrrrrcrrrrrrrrrcrrrrrrrr@{}} & \phantom{rrrrrrrrr}& \\
\midrule 
Control System/\\Fault type & Fault-free & Bump left & Bump right & Flat left & Flat right  & Rank\\ \midrule
\textit{ $x_e$ RMSE}\\
ROBUST & 0.0156 & 1.3897 & 1.7950 & 0.1430 & 0.1434 & 3\\
ADAPTIVE & 0.0036 & 1.3777 & 1.7830 & 0.1310 & 0.1314 & 2\\
RESILIENT & 0.6948 & 5.5747 & 5.1694 & 6.8214 & 6.8219  & 4\\
ANTIFRAGILE  & 0.0025 & 1.3766 & 1.7819 & 0.1299 & 0.1300 & 1\\
\midrule
\textit{$y_e$ RMSE}\\
ROBUST & 0.0005 & 0.3538 & 1.2162 & 0.1689 & 0.6052 & 2\\
ADAPTIVE & 0.0007 & 0.3540 & 1.2159 & 0.1691 & 0.6053 & 3\\
RESILIENT & 0.0316 & 0.3848 & 1.2851 & 0.1999 & 0.6362 & 4 \\
ANTIFRAGILE  & 0.0002 & 0.3529 & 1.2170 & 0.1681 & 0.6043 & 1\\
\midrule
\textit{$\phi_e$ RMSE}\\
ROBUST & 0.05521 & 0.2573 & 0.3441 & 0.2332 & 0.8642 & 1\\
ADAPTIVE & 0.0917 & 0.7628 & 0.0807 & 0.6785 & 1.3898 & 4\\
RESILIENT & 0.5569 & 0.2976 & 0.3844 & 0.2133 & 0.8945  & 2\\
ANTIFRAGILE  & 0.0707 & 0.1838 & 0.1016 & 0.6195 & 1.3807 & 3\\
\bottomrule
\end{tabular}
\caption{Performance evaluation for the different trajectory control algorithms in fault-free operation and under the impact of 4 types of faults (i.e. 2 sensor faults --modeled as flat tires-- and 2 actuator faults --modeled as a motor shaft periodic bump-- of robot's driving wheels). For the quantitative evaluation, we consider the RMSE-based ranking of minimal robot position deviations (i.e. minimizing all/most of errors, $x_e$, $y_e$, and $\phi_e$) with rank as the ordered average RMSE in faulty and fault-free operation. A lower ranking order is better.}
\label{perf-table}
\end{table*} 

As we can see, ANTIFRAGILE control dominates the position control of the robot in the presence of uncertainties, followed closely by the ROBUST and ADAPTIVE control strategies, and lastly by RESILIENT control. Interestingly, ANTIFRAGILE control places only third when it comes to controlling the heading of the robot, where ROBUST control and RESILIENT control excel due to explicit decoupling of the heading from the Cartesian positioning in the control law design. There are also more subtle aspects that we will unfold in the following section.

We remind the reader that the framework we developed along with the different controllers is available in the codebase on \href{https://gitlab.com/akii-microlab/antifragile-robot-control/}{GitHub}\footnote{Codebase available at \href{https://gitlab.com/akii-microlab/antifragile-robot-control/}{https://gitlab.com/akii-microlab/antifragile-robot-control/}}. Using the codebase one can explore and test arbitrary hypotheses on the closed-loop control behavior in the presence of single faults, cascaded faults, or other parametrized uncertainty types. This possibility extends the initial exploration we performed in the present manuscript and offers the users more interesting possibilities to investigate the benefits of antifragile robot control.

\begin{figure}[t!]
\includegraphics[scale=0.35]{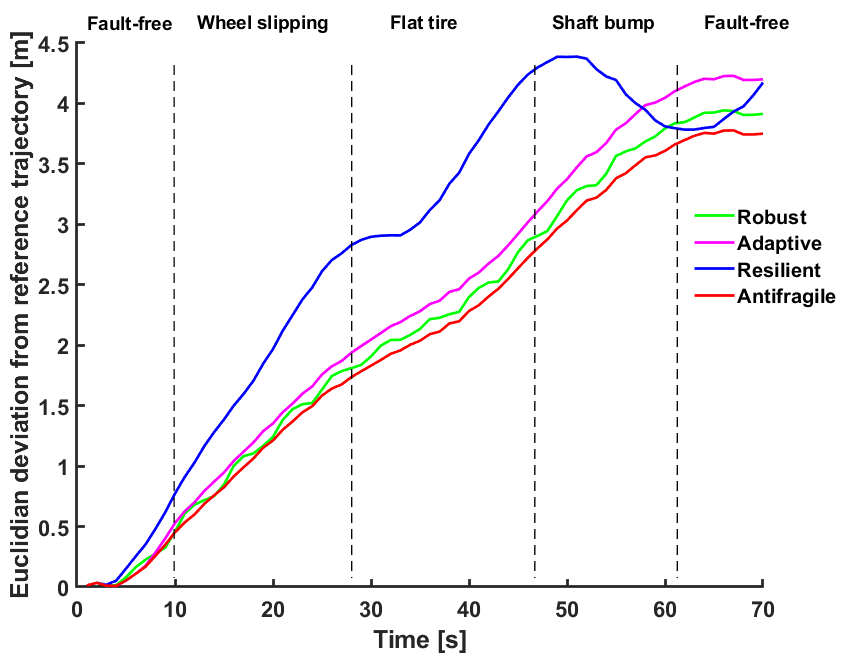}
\centering
\caption{Trajectory tracking performance analysis in the presence of cascaded faults. After starting the operation with no faults (i.e. fault-free region) the faults and uncertainty injection system introduces one after the other the three types of faults at different times (i.e. $t_{slipping} = 10s$, $t_{flat} = 28s$, $t_{bump} = 46s$, and back to fault-free from $t_{free} = 62s$ onward).}
\label{fig13}
\end{figure}

The final experiment we performed in this study comes back to the fragility-robustness-antifragility spectrum in Figure~\ref{fig1}. More precisely, we wanted to evaluate the performance of the four control strategies on cascaded faults that occur at random moments during the robot's operation and with different magnitudes. We defined a scenario where we subsequently injected: changes in environment parameters (e.g. wheel slippage), sensor faults (e.g. perceiving a continuous wheel radius decrease during operation akin to a flat tire), and actuator faults (e.g. a periodic eccentric mechanical motion of the DC motor shaft akin to a wheel bump), respectively. We then compared the four strategies in terms of the tracking performance (i.e. RMSE on $x_e$, $y_e$, and $\phi_e$) under the effect of the faults. Our findings are depicted in Figure~\ref{fig13}, where we can definitely see the superior performance of the ANTIFRAGILE and ROBUST controllers which overcome the ADAPTIVE control and RESILIENT control. Interestingly, the diagram still captures the layout of the fragility-robustness-antifragility spectrum, where RESILIENT control has the loosest reaction (i.e. slow) to the occurrence of faults, but demonstrates in between a good stationary behavior (i.e. see the blue trace in Figure~\ref{fig13} between $25 s$ and $33 s$ where the position error doesn't increase, hence the fault was accommodated by the controller. In the case of ANTIFRAGILE and ROBUST, we can detect high oscillations due to the variable structure control law which keeps the position error as low as possible with the price of a high control activity. The ADAPTIVE controller manages to provide stable performance under the cascaded faults with comparable performance with the ANTIFRAGILE and ROBUST strategies. 

After introducing the experimental setup and results, we now turn to a more in-depth analysis of the results in the following section. 

\section{Discussion}

Trajectory tracking is a fundamental problem in mobile robot control and an even more fundamental control issue when considering uncertainty and sensor and actuator faults. Control strategies designed for tackling this problem need to synthesize control laws for the robot's actuators which compensate for longitudinal $x_e$, lateral $y_e$, and heading angle $\phi_e$ errors under uncertainty in both the operating environment and robot's sensors and actuators reliability. 
In our study, we introduce a novel control strategy termed ANTIFRAGILE control, which has the benefit of gaining from continuous exposure to uncertainty and reaching performance that is beyond robust. We have validated our hypothesis (depicted in Figure~\ref{fig1}) through a batch of experiments, an extensive evaluation, and the design of a framework for the evaluation of fault and uncertainty injection in mobile robot trajectory tracking control. 
The competitive algorithms were chosen among the state-of-the-art approaches for trajectory tracking, namely ROBUST control (i.e. a sliding mode controller), ADAPTIVE control (i.e. model predictive controller), and RESILIENT control (i.e. fuzzy logic controller). The experiments and evaluation were designed to capture the comparative performance degradation of the closed-loop controllers in the presence of sensor and actuator faults.

\subsection{Time scale separation}

A strong component in the ANTFRAGILE control synthesis is the time scale separation component responsible for the low-level actuator control robustness. Using singular perturbation theory, we have implemented time scale separation within the ANTFRAGILE controller based on the analysis of the dynamics boundary layers shapes (see Figure~\ref{fig6}) and the shape of the prescribed path and curvature quantities given as a reference for the actuators. Interestingly, we could obtain a separation of the control regimes in antifragile and fragile based solely on the curvature of the uniform solution shape of the closed-loop system, given by Equation~\ref{curvature}. This separation is then exploited in the computation of the actuator control signal (i.e. $v(t)$ and $w(t)$) which takes either the form of a "rapid" or "slow" dynamics.
A similar behavior, or at least comparable, is achieved in the ROBUST design through the variable structure control. More precisely, the lower-order design of the controller using equivalent control (see Equation~\ref{robustctrl}) accounts for a reduced-order technique analogous to the effect singular perturbation offers. The ADAPTIVE control attempts a time scale separation through the choice of a multi-scale cost function with regularizing terms for each temporal scale magnitude (see Equation~\ref{adaptivectrl}). This is especially visible in the MPC instantiation we considered where the prediction horizon can be weighted separately on "fast" and "slow" dynamics. 
Finally, the RESILIENT control can induce, in principle, such time scale separation explicitly. In our case, this can be achieved through fuzzy inference rules that capture the co-variance of the first derivative of error terms and their rate of change (see Equation~\ref{resilientctrl}). Time scale separation is a design component that determines the low-level actuator control and the benefits of a curvature-aware synthesis of control law (see Figure~\ref{fig6}).

\subsection{Redundant over compensation}

Redundant over compensation refers to the capacity of the controller to build capacity in compensating (in a timely manner) for uncertainty and faults. This "capacity" building can be seen as a measure of compensation, which goes beyond accommodating the uncertain event and up to gaining (i.e. sudden convergence of error) from the unexpected event. ANTIFRAGILE control uses redundant over compensation when designing the low-level control of the actuators, depicted in Figure~\ref{fig7}. After identifying the "fast" and the "slow" dynamics of the closed-loop system of the actuators, the design focuses on rewriting the dynamics such that the closed-loop system dynamics are described solely by the solution in the overlap region in Figure~\ref{fig6}, where convexity of the response can be probed through Equation~\ref{curvature}.
In order to analyze the redundant over compensation behavior of the competitive control strategies, we start with a thorough overview of the experimental results in Figure~\ref{fig9}. Here, we can observe the differences in compensating for the curvature (i.e. second order effects) of the prescribed dynamics (i.e. the trajectory is a place and a dynamics). More precisely, in Figure~\ref{fig9} b we observe how the ANTIFRAGILE and ROBUST controllers follow the prescribed trajectory closely (see Table~\ref{perf-table} for quantitative assessment), with small magnitude overshooting in high-curvature regions. On the other side, the ADAPTIVE and RESILIENT controllers capture the overall inflections of the trajectory but fail to smoothly capture highly convex regions and the end position.
Finally, to get a more intuitive understanding of the benefits of redundant over compensation, we analyze the results in Figure~\ref{fig10}. Here, the kinematic (i.e. linear and angular velocities) and the dynamic (i.e. linear and angular acceleration) quantities describing the robot's motion are analyzed, with respect to the reference, control, and real velocities and accelerations. A trademark of ANTIFRAGILE control is the fact that the control linear velocity signal overshoots at regions where the curvature sign changes (see Figure~\ref{fig10} a) on the trajectory, visible also in the rate of change of velocity, depicted in Figure~\ref{fig10} b. These high-frequency changes are also determined by the variable structure control synthesis at the core of the ANTIFRAGILE design. This "capacity" building is also visible in the angular control signals, where both angular velocity control signals, depicted in Figure~\ref{fig10} c, and their rate of change surpasses shortly the prescribed values at the high-curvature inflections of the trajectory. This behavior is clearly motivated by the simplified dynamics in Equation~\ref{innerctrlsingpertctrl} which basically describe a proportional effect to changes in the dynamics of the "fast variables" (see Equation~\ref{dcmotorsingpert}).

\subsection{Variable structure control}

The final ingredient of ANTIFRAGILE control design is the variable structure control synthesis. This approach is highly used in the realm of robust control design as a means to inherently handle uncertainty, be it structured (i.e. parametric uncertainty) or unstructured (i.e. unmodelled dynamics). This is also the common design component between the ROBUST and ANTIFRAGILE controllers. Such a control pushes the system to a manifold that describes the prescribed dynamics of the closed-loop system and ensures that the system stays there. As mentioned earlier, the manifold becomes a place and a dynamics, as depicted in Figure~\ref{fig8}. Intuitively, the control signal to generate will be discontinuous in nature and stability is a strong prerequisite (see the analysis in Section 2.2 on control design and Equations~\ref{controllawacc},~\ref{controllawomega}, and ~\ref{lyapunovvelocityfinal}, respectively). 
The induced behavior of the variable structure control in both the ROBUST and ANTIFRAGILE controllers is visible in Figure~\ref{fig10}. This is even more clear when analyzing the performance in Table~\ref{perf-table}. Here we can see when only considering the fault-free (baseline) scenario, that the ANTIFRAGILE and ROBUST controllers excel in providing minimal RMSE on longitudinal and lateral deviations, which overcome both the ADAPTIVE and RESILIENT controllers. The dominance is changed in the heading error, where ANTIFRAGILE only ranks three due to its implicit weighting of the heading in the manifold design (please refer to Equation~\ref{surfacedesign}). This is further emphasized in Figure~\ref{fig11} and motivated by the fact that in trajectory tracking heading is secondary whereas the overall (Euclidean) position needs to match as good as possible the prescribed trajectory.
Due to the underlying model predictive control, the ADAPTIVE control does a comparatively good job across deviations RMSE in the fault-free scenarios, even better than the RESILIENT control which excels in the heading error minimization.
When considering the scenarios with uncertainty and faults, we considered a performance evaluation for the different trajectory control algorithms under the impact of 4 types of faults (i.e. 2 sensor faults --modeled as flat tires-- and 2 actuator faults --modeled as a motor shaft periodic bump-- of robot's driving wheels), as shown in Table~\ref{perf-table}. Overall, but with a rather minimal margin from ROBUST, the ANTIFRAGILE control dominates the other control strategies with minimal RMSE across all fault types. Closely, the ROBUST control excels in orientation error minimization, outperforming ANTIFRAGILE and the other strategies. ADAPTIVE control comes close to ROBUST with a small penalty that might be based on the choice of the cost function. Finally, RESILIENT control provides a more slow varying response (akin to the hypothetical situation in Figure~\ref{fig1}) but with a stable outcome. 
Finally, in our last and most extreme example, we cascaded faults and uncertainty in the robot's trajectory tracking operation (see Figure~\ref{fig13}). The overall evaluation criteria were chosen for the Euclidean deviation from the prescribed trajectory. As one can see, and also supported by the previous discussion and analysis, the experiments bring us closer to validating the hypothesis (visually described in Figure~\ref{fig1}).
The analysis in Figure~\ref{fig13} shows that ANTIFRAGILE control (red trace) offers the smallest deviation with small regions (typically before a new fault occurs) where the errors actually decrease even more. Following closely is the robust behavior of the ROBUST controller which, given its variable structure control law, exhibits a high-frequency oscillatory control law determining oscillations in the actuators commands and subsequently in the trajectory (see the green trace in Figure~\ref{fig13}). ADAPTIVE exploits the advantages of MPC and provides good performance by exploiting the predictive nature of the underlying model and receding horizon. Finally, resilient slowly reacts to each injected fault but accommodates after a transient fault but with the price of a higher overall position error.

As our experiments show, ANTIFRAGILE control has the potential to offer beyond ROBUST performance in the presence of uncertainty, sensor, and actuator faults. This is very useful in applications such as the ones we described in our preamble, where comfort is an important dimension of the task. We believe, that such an ANTIFRAGILE control design can provide an interesting path towards the closed-loop system which gains from uncertainty, a goal long sought in autonomous robotics.

\section{Conclusion}

Modeling and handling uncertainty in closed-loop robot control tasks is still an openly debated and fruitful area of research. In an arena where control theory provides its most powerful tools and robotics provides its more pragmatic deployments, emerging approaches need to overcome well-established "recipes". ANTIFRAGILE control is a new approach to control, which approaches control synthesis from the perspective of capturing the peculiarities of the response of the system to control. First and second-order effects provide useful hints on where and how to issue control signals that can drive the systems in regions of the solutions space where the system is not only robust to uncertainty and volatility but can also gain from it and anticipate future uncertain events. This is the core motivation of ANTIFRAGILE control. 
The current study is an exploratory one, along with the previous instantiations of ANTIFRAGILE control in \cite{axenie2022antifragile} and \cite{axenie2022antifragiletraffic}, and is meant to "instigate" the community to adopt and leverage a novel control system design where crucial design information lies in metrics of the shape of the system response to uncertainty.
In the current instantiation of ANTIFRAGILE control for mobile robots' trajectory tracking, we have only scratched the surface of the possibilities such a framework offers. The experiments with parametrizable faults helped us validate the framework and the controller design for a relatively simple task and dynamics. We are keen to build a consistent thesis and framework around the principles of ANTIFRAGILE control and open the path for induced antifragility in technical systems.

\end{document}